\documentclass[letterpaper, 10 pt, conference]{ieeeconf}
\IEEEoverridecommandlockouts
\usepackage{amsmath,amsfonts}
\usepackage{algorithmic}
\usepackage{algorithm}
\usepackage{array}
\usepackage[caption=false,font=normalsize,labelfont=sf,textfont=sf]{subfig}
\usepackage{textcomp}
\usepackage{stfloats}
\usepackage{url}
\usepackage{verbatim}
\usepackage{graphicx}
\usepackage{xspace}
\usepackage{booktabs}
\usepackage{multirow}
\usepackage{colortbl}
\usepackage{xcolor}
\usepackage{balance}
\usepackage[colorlinks=true, allcolors=blue]{hyperref}
\usepackage{glossaries}

\newcommand{\objnav}{\textsc{ObjectNav}\xspace}
\newcommand{\semnav}{\textsc{SemNav}\xspace}

\newcommand{\moveforward}{\textsc{move\_forward}\xspace}
\newcommand{\movebackward}{\textsc{move\_backward}\xspace}
\newcommand{\turnleft}{\textsc{turn\_left}\xspace}
\newcommand{\turnright}{\textsc{turn\_right}\xspace}
\newcommand{\stopac}{\textsc{stop}\xspace}
\DeclareMathOperator*{\argmin}{arg\,min}

\def\BibTeX{{\rm B\kern-.05em{\sc i\kern-.025em b}\kern-.08em
    T\kern-.1667em\lower.7ex\hbox{E}\kern-.125emX}}

\title{\LARGE \bf
\semnav: Enhancing Visual Semantic Navigation in Robotics through Semantic Segmentation
}

\author{Rafael Flor-Rodríguez-Rabadán$^{1}$, Carlos Gutiérrez-Álvarez$^{1}$, Francisco Javier Acevedo-Rodríguez$^{1}$,\\ Sergio Lafuente-Arroyo$^{1}$ and Roberto Javier López-Sastre$^{1}$
\thanks{This work was supported by projects: GUIDANCE-4D, with reference CM/DEMG/2024-028, of CAM-UAH; NAVIGATOR-D, with reference PID2023-148310OB-I00 from the Ministry of Science and Innovation of Spain; and MOEVE Chair at the University of Alcalá.}
\thanks{$^{1}$University of Alcalá, Department of Signal Theory and Communications, Alcalá de Henares, 28805, Spain.
        {\tt\small rafael.flor@uah.es}}
}

\begin{document}

\maketitle
\thispagestyle{empty}
\pagestyle{empty}

\begin{abstract}
Visual Semantic Navigation (VSN) is a fundamental problem in robotics, where an agent must navigate toward a target object in an unknown environment, mainly using visual information.
Most state-of-the-art VSN models are trained in simulation environments, where rendered scenes of the real world are used, at best.
These approaches typically rely on raw RGB data from the virtual scenes, which limits their ability to generalize to real-world environments due to domain adaptation issues.
To tackle this problem, in this work, we propose \semnav, a novel approach that leverages semantic segmentation as the main visual input representation of the environment to enhance the agent’s perception and decision-making capabilities.
By explicitly incorporating this type of high-level semantic information, our model learns robust navigation policies that improve generalization across unseen environments, both in simulated and real world settings.
We also introduce the \semnav dataset, a newly curated dataset designed for training semantic segmentation-aware navigation models like \semnav.
Our approach is evaluated extensively in both simulated environments and with real-world robotic platforms.
Experimental results demonstrate that \semnav outperforms existing state-of-the-art VSN models, achieving higher success rates in the Habitat 2.0 simulation environment, using the HM3D dataset.
Furthermore, our real-world experiments highlight the effectiveness of semantic segmentation in mitigating the sim-to-real gap, making our model a promising solution for practical VSN-based robotic applications.
The code and datasets are accessible at \url{https://github.com/gramuah/semnav}.
\end{abstract}

\section{Introduction}
%Motivation of the problem
Autonomous navigation remains a fundamental challenge in robotics, particularly in unstructured and dynamic environments.
Traditional approaches, such as the ones using Simultaneous Localization and Mapping (SLAM), for example, focus on a geometric reconstruction of the environment in conjunction with path planning and obstacle avoidance techniques (e.g. \cite{slam,rosinol2020}).
However, these classical models typically struggle when they need to generalize across environments and domains~\cite{cadena2016}.

Visual Semantic Navigation (VSN) has emerged as an alternative paradigm that leverages advances in machine learning and the availability of large-scale data.
VSN models are essentially learning-based navigation approaches that apply end-to-end methods to directly map sensory visual inputs to actions that control the robot (e.g.~\cite{ramrakhya2022,yadav2022, chaplot2020,chang2020,wijmans2020,YokoyamaHM3DOVONAD,ye2021,yang2018,GutierrezAlvarez2023VisualSN}).

\begin{figure}
    \centering
    \includegraphics[width=0.8\linewidth]{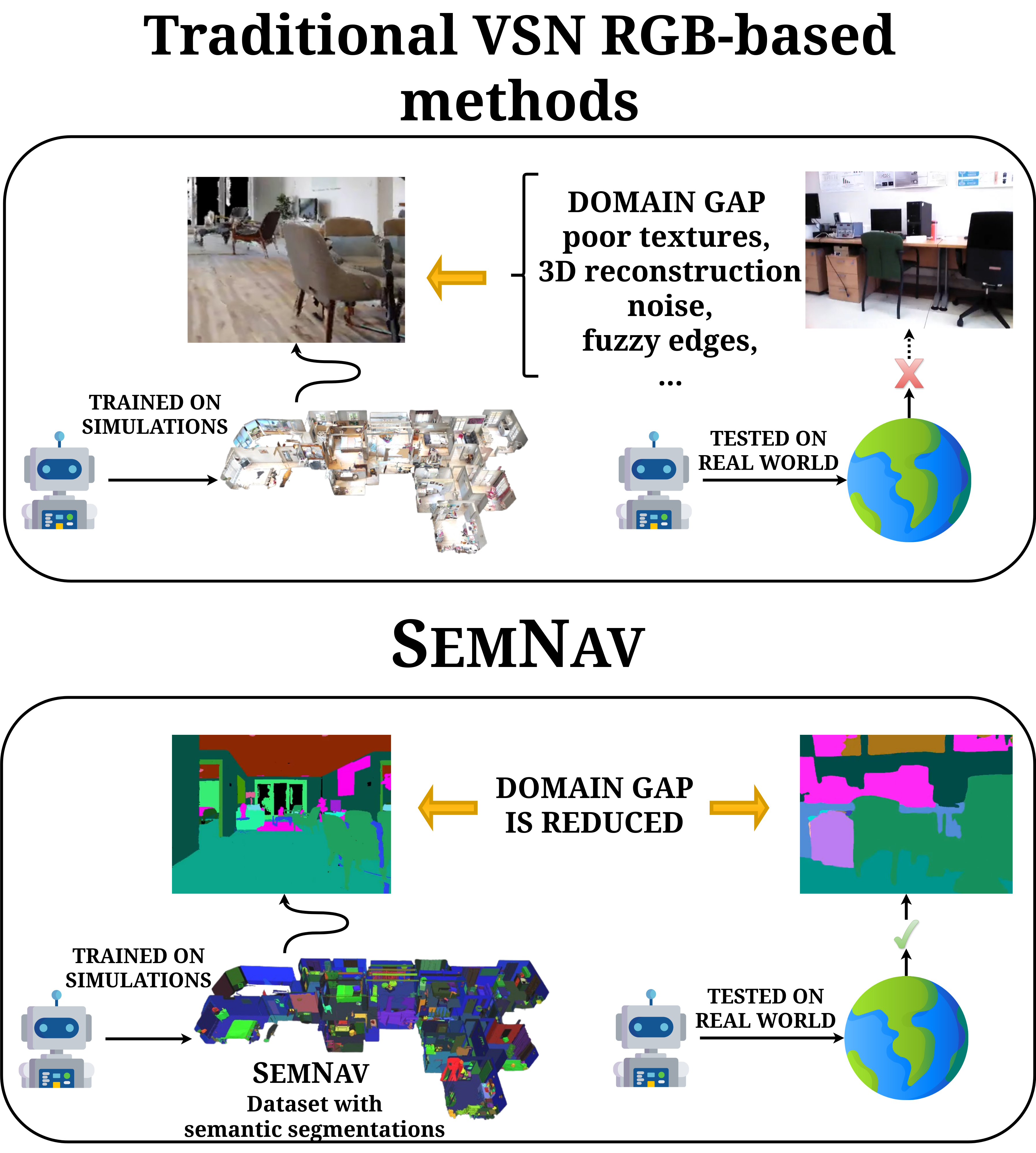}
    \caption{Traditional VSN models are typically trained in simulated environments, relying primarily on RGB images. However, the substantial domain gap between simulated and real-world imagery often leads to poor performance when these systems are deployed on real robots. To overcome this limitation, we introduce the \semnav paradigm, where VSN systems use semantic segmentation maps of the environment as their main sensory input. This approach significantly reduces the domain gap between training simulations and real-world scenarios, enabling models to generalize more effectively and perform reliably when tested outside of simulation.}
    \label{fig:ga}
\end{figure}

VSN approaches, instead of planning a geometric route over a map, e.g. \cite{werby2024hierarchical,Cai2023DGMemLV,Gu2023ConceptGraphsO3},
directly learn navigation policies that enable robots to explore unknown environments while following semantically meaningful instructions, such as ``Move towards a chair'' or ``Find a bedroom''.
This particular problem is known as the Object Navigation (\objnav) problem~\cite{habitatchallenge2023}, where an agent must navigate, in an unknown scenario, so without any map, toward an instance of a given object category using mainly visual information.
All these VSN \objnav systems are trained and evaluated in complex simulation platforms, such as Habitat~
\cite{NEURIPS2021_021bbc7e, habitatchallenge2023}, RoboTHOR~\cite{robothor} or ProcTHOR~\cite{procthor}, where the
agents learn to navigate in interactive 3D environments and complex physics-enabled scenarios.
AI2-THOR and ProcTHOR enable interaction with realistic virtual 3D environments generated through computer graphics.
Habitat, on the other hand, allows agents to interact with 3D models corresponding to real-world 3D scans of scenes
and houses, such as those included in the HM3D datasets~\cite{Ramakrishnan2021HabitatMatterport3D,yadav2022habitat}.
VSN-based approaches have practical real-world applications, such as the development of assistive robots capable of retrieving objects and providing support to individuals in need.
These models operate in a plug-and-play manner, enabling autonomous and intelligent exploration of the environment without requiring prior map construction~\cite{savva2019habitat}.
Furthermore, VSN systems can be used for autonomous exploration of unseen or partially known environments, facilitating tasks such as search and rescue, environment monitoring, and efficient scene understanding in dynamic or unstructured spaces.

Despite recent progress, state-of-the-art \objnav models often struggle with real-world generalization due to the significant domain gap between simulation-trained policies and real-world execution~\cite{gervet2022, Wasserman2023ExploitationGuidedEF, GutierrezAlvarez2023VisualSN}.
In other words, no matter how realistic the simulation environments used during learning are, the real world presents other challenges.
To begin with, the images provided by the simulation platforms are either not entirely realistic or exhibit artifacts typical of 3D scanning processes of real-world scenes.
The egocentric images that a robotic platform will acquire in the real world will present higher levels of blurring due to the platform's movement, more diverse lighting situations, and will also include changes that naturally occur in dynamic environments.
This results in navigation strategies that fail to generalize beyond the training and simulated domain.
To address this challenge, we propose \semnav (see figure \ref{fig:ga}), a novel \objnav approach that uses semantic segmentation as the main visual input.
Our goal in basing the visual perception of the VSN system on semantic segmentation is twofold.
First, we argue that a representation based on semantic segmentation naturally mitigates the domain adaptation problem between the real world and simulated environments.
The differences between the semantic segmentations of a real-world scene and those of a rendered scene in a simulation environment such as Habitat~\cite{NEURIPS2021_021bbc7e} are smaller than those found in RGB images from both domains.
Second, by incorporating semantic priors into the visual representation of the environment, our model learns robust navigation policies that better adapt to unseen environments, hence improving the success of the navigation task.
Our approach is inspired by the intuition that human navigation heavily relies on semantic cues—understanding that a ``kitchen'' typically contains a ``sink'' or a ``refrigerator'' enables more efficient exploration.

%Contributions
To validate our approach, we conduct extensive experiments in both simulated and real-world environments.
Our findings indicate that semantic segmentation significantly improves navigation performance, particularly in bridging the sim-to-real gap.
Unlike conventional methods that degrade in real-world deployments, \semnav exhibits higher robustness by leveraging structured semantic information.
Overall, our main contributions in this work are as follows:
\begin{itemize}
    \item We release the new \semnav dataset designed for training semantic segmentation-aware navigation models, enabling further research in this domain.
    \item We present \semnav, a novel \objnav model designed to employ the semantic segmentation of the robot's egocentric view for learning the mapping between visual observations and navigation actions in unexplored environments. \semnav is versatile, demonstrating robust and efficient navigation and exploration behaviors in both simulated environments and real-world scenarios.
    \item In our experimental evaluation, the \semnav approach demonstrates superior performance compared to state-of-the-art methods, both on the Habitat 2.0 simulation platform~\cite{habitatchallenge2023} and in real-world testing.
\end{itemize}

\section{Related work}
\label{sec:relatedwork}
Robotic autonomous navigation has long been a key focus in research, with many methods developed to tackle the challenge of maneuvering robots through complex and dynamic environments.
\textbf{Classical navigation} approaches are mainly based on Simultaneous Localization and Mapping (SLAM)~
\cite{slam,Xu2025DeepLV,rosinol2020,campos2021,SLAM++}.
SLAM-based solutions primarily tackle the navigation problem by focusing on obstacle avoidance, map creation and path-planning algorithms.
However, these models face challenges in generalizing to different environments~\cite{cadena2016}.

To address these limitations and capitalize on recent breakthroughs in machine learning and the availability of large-scale datasets, research has increasingly shifted toward \textbf{learning-based} approaches for robotic navigation.
Within this category, Visual Semantic Navigation (VSN) models stand out.
Unlike traditional methods that depend on geometric path planning over pre-constructed maps, VSN models adopt a fundamentally different strategy by learning navigation policies that enable robots to navigate unfamiliar environments based on semantically meaningful instructions, such as ``Go to a chair'' (\objnav problem) or ``Find this image'' (\textsc{ImageNav} task).
For a comprehensive overview of recent progress and challenges in VSN, interested readers may refer to~\cite{habitatchallenge2023}.
Fundamentally, VSN models learn to make navigation decisions based mainly on visual information (RGB and/or depth data)~\cite{ramrakhya2022,yadav2022, chaplot2020,chang2020,wijmans2020,YokoyamaHM3DOVONAD,ye2021,yang2018}.
Two main VSN training approaches include both Imitation Learning (IL) and Reinforcement Learning (RL).
IL-based methods learn navigation policies from previously annotated demonstrations~\cite{ramrakhya2022,yadav2022}.
RL-based methods can be split into two categories.
The first category consists of models that follow an end-to-end RL approach~\cite{zhu2017, wijmans2020, YokoyamaHM3DOVONAD} via interaction with the environment.
Several training strategies have been proposed for these systems, including the use of auxiliary tasks~\cite{ye2021}, object-relationship graphs~\cite{yang2018}, and the fusion of visual and auditory information~\cite{Kondoh2023MultigoalAN}.
The second category includes works that employ modular learning approaches~\cite{chang2020,gervet2022, Kang2024HSPNavHS, Wasserman2023ExploitationGuidedEF}, where the navigation process is divided into independent modules addressing different aspects of navigation.

%Other recent alternatives
In recent years, advances in Large Language Models (LLMs) have enabled their application to the visual semantic navigation problem, giving rise to \textbf{Vision-Language Navigation} (VLN) models~\cite{He2023MLANetMA,Huang2022KnowledgeDP,Wang2025}.
In these models, an embodied agent executes natural language instructions within real 3D environments to navigate.
Finally, we have the recent \textbf{diffusion-based navigation} approach where the latest advancements in generative modeling, particularly in diffusion models \cite{Ho2020,Rombach2022} have influenced robot navigation~\cite{vint,FlowNav}.

%Our main contributions
As previously argued, despite significant efforts in developing VSN models, a major limitation persists: most models are trained and evaluated predominantly in simulation environments.
In fact, only a few studies have attempted to evaluate VSN solutions in real-world settings~\cite{gervet2022, Wasserman2023ExploitationGuidedEF,GutierrezAlvarez2023VisualSN}.
The results reported in these studies reveal a significant domain gap problem: while VSN models achieve high success rates in simulation, their performance deteriorates drastically when deployed in real-world environments.
In this work, we propose a novel VSN model, named \semnav, designed to learn navigation policies for the \objnav problem using a semantic segmentation of the robot's egocentric view.
The semantic segmentation output is fed into a Convolutional Neural Network (CNN)-based visual encoder, which extracts navigational cues from the segmentation information to enhance navigation in unseen environments.
To the best of our knowledge, no previous VSN work has implemented this idea as we have.
Some studies leverage semantic segmentation to define navigable regions and plan routes~\cite{Adachi2019,Adachi2023}, while others combine this information with object detection systems~\cite{Mousavian2019} to enable target-driven visual navigation.
Our initial hypothesis, confirmed by experimental results, is that semantic segmentation serves as a valuable source of information that allows: (1) the development of more efficient VSN models and (2) the natural mitigation of domain adaptation issues.

\section{\semnav Approach}
\label{sec:semnav}

\subsection{\semnav dataset}
\label{sec:dataset_semnav}
%Novel dataset
The first essential requirement for training a VSN model that utilizes semantic segmentation as its visual input is the availability of a dataset containing this type of information.
In this work, we extend the HM3D dataset~\cite{ramrakhya2022} to create the \semnav dataset, which introduces support for a semantic segmentation sensor.
Specifically, we detail the enrichment process that enables the generation of a semantic segmentation sensor compatible with the Habitat simulator~\cite{NEURIPS2021_021bbc7e}.
This semantic sensor allows any agent interacting with Habitat to access an egocentric view that corresponds to the semantic segmentation of the scene currently perceived by the robot.

\begin{figure}
    \centering
    \includegraphics[width=0.7\linewidth]{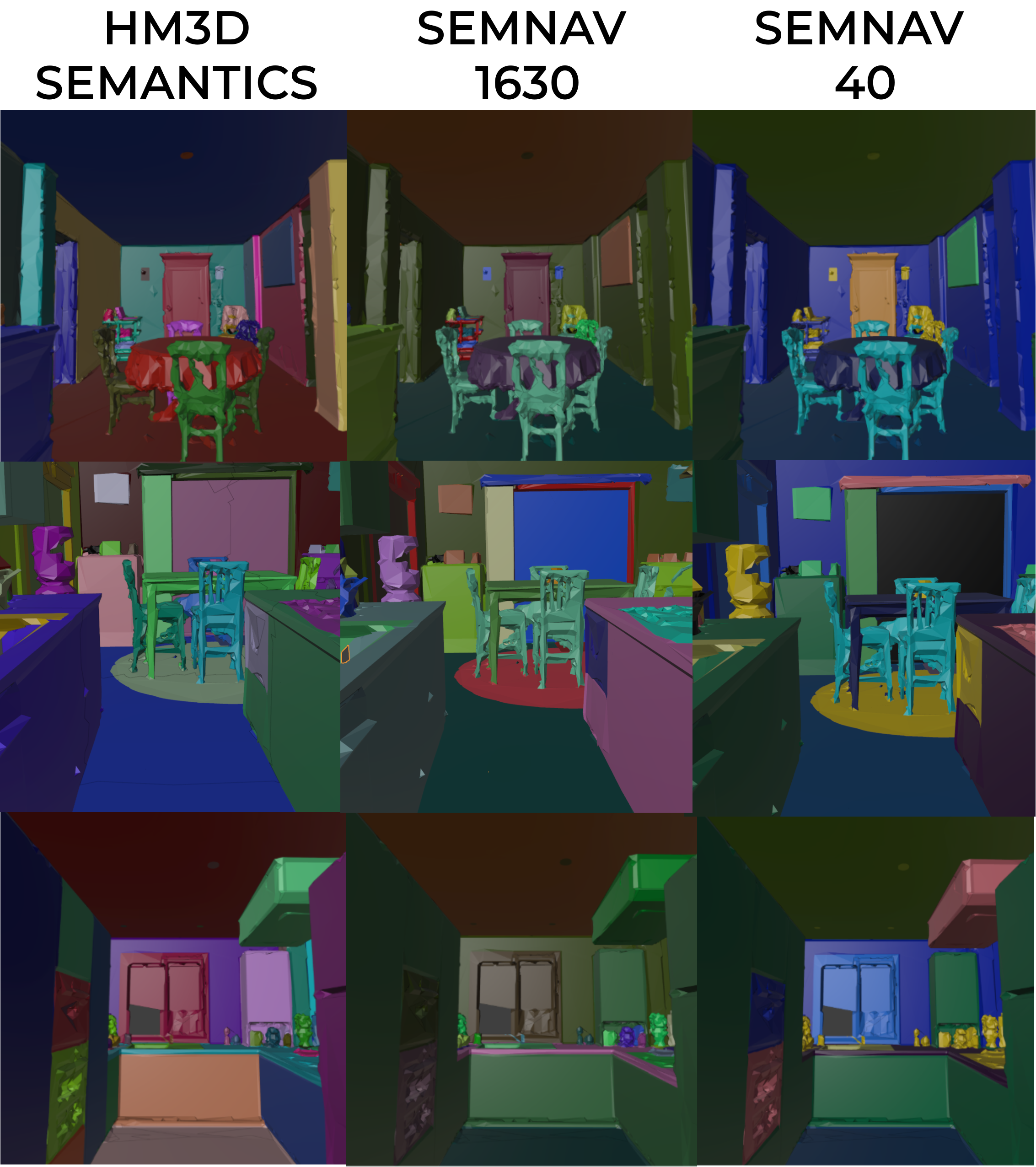}
    \caption{Comparison between the HM3D dataset and the \semnav 1630 and \semnav 40 novel datasets.
    In HM3D, objects like chairs are inconsistently labeled with different colors across and within scenes. In contrast, \semnav 1630 assigns a uniform color to objects of the same category while reflecting finer distinctions, such as differentiating kitchen and dining tables. \semnav 40 further simplifies the labeling by merging less critical categories, grouping both types of tables under a shared label to streamline navigation tasks. Figure is best viewed in colour.}
    \label{fig:datasets}
\end{figure}

The HM3D dataset provides a collection of 3D real-world spaces, densely and manually annotated with semantic information, as illustrated in Figure \ref{fig:datasets}.
This dataset comprises over 140,000 object instance annotations distributed across 216 3D environments and approximately 3,100 rooms.
The semantic data is embedded in the form of texture images mapped onto the original 3D geometry of the HM3D scenes.
This information is packed into binary glTF format files, one per scene.
Each object instance is assigned a unique color identifier, which maps to a corresponding textual label specifying the object’s category, stored in a separate annotation file.

The separation of object texture information and semantic labels into different files leads to a limitation: the built-in sensor in Habitat for the HM3D dataset does not deliver a \emph{true} semantic segmentation of the scene.
For instance, chairs, as illustrated in Figure \ref{fig:datasets}, are not consistently assigned the same identifier or label across different scenes, but treated as different textures.

Our model, \semnav, relies on semantic segmentation to navigate effectively.
To meet this requirement, we have developed an enhanced semantic segmentation sensor.
It builds upon the original semantic information available in the HM3D dataset.
Unlike the default information, our sensor ensures that object categories are consistently assigned unique and uniform labels across all scenes.
Technically, the creation of this sensor has been automated by designing a mapping process that allows translating the information associated with textures into semantic segmentation labels that are now expanded both intra- and inter-scenes.
This improvement guarantees that different instances of the same object category receive identical labels within a single scene and across multiple scenes, as illustrated in Figure~\ref{fig:datasets}.
Using the semantic annotation files (one per scene), it is possible to determine the semantic category associated with each object.
To generate our sensor, we used these annotations to convert the packed texture information in the 3D model files into semantic segmentation data.
We assigned a unique global ID and color to each annotation and replaced the original textures with their corresponding semantic segmentation colors across all scenes and rooms.
After this conversion, we integrated the semantic segmentation sensor into the simulator, allowing any agent to query the simulator for the semantic segmentation of its egocentric view in the rendered 3D environment.

To provide flexibility and address varying levels of semantic granularity, we have created two semantic segmentation sensors: one with 1,630 distinct object categories, referred to as \semnav 1630, and another with a more compact set of 40 categories, called \semnav 40.
The \semnav 1630 sensor was generated by leveraging the fine-grained object annotations available in HM3D~\cite{ramrakhya2022}, where each manually labeled object instance was assigned to a unique category.
This allowed identically annotated objects to be grouped under the same category.
However, upon conducting a detailed analysis, we identified certain limitations in these annotations, including noise from overly specific object labels and occasional misclassifications.
To address these challenges and improve the robustness of the semantic segmentation, we created a second semantic segmentation sensor, \semnav 40.
In this version, each of the 1,630 original annotations was manually mapped to one of the 40 broader categories defined by the NYUv2 dataset~\cite{Silberman2012IndoorSA}, a widely recognized benchmark for indoor semantic segmentation.
This mapping is provided as metadata within the HM3D dataset~\cite{ramrakhya2022}, which we directly used in our mapping process, for the generation of the novel semantic segmentation sensor.
This reduction from fine-grained labels to 40 coherent categories simplifies navigation and minimizes the impact of noisy annotations.
Both \semnav 1630 and \semnav 40 sensors were integrated into the Habitat simulator, allowing any agent to query the simulator for a true semantic segmentation corresponding to its egocentric view in the 3D environment being rendered.

From this point onward, we will refer to each sensor as if it were a dataset in its own right.
Accordingly, we have released two datasets, \semnav 1630 and \semnav 40, which are fully integrated into the Habitat platform.
This integration enables straightforward training of any model requiring semantic segmentation input, streamlining the development of advanced navigation models and reducing implementation overhead.
Both datasets are accessible at \url{https://github.com/gramuah/semnav}.

\subsection{\semnav model}
\label{sec:model_semnav}

In this work, we also introduce the \semnav model, designed to address the VSN problem known as \objnav~\cite{habitatchallenge2023}.
The objective of an \objnav navigation episode is to enable an agent to navigate in a scene $S_i$ from a set of available scenes $\mathcal{S} = \{S_1, \dots, S_n\}$, towards an object of a specific category $c_i$ belonging to the category set $\mathcal{C} = \{c_1, \dots, c_m\}$, starting from an initial position $p_0$ in the navigation environment.
The agent must reach an instance of the target category and execute a stop action before exceeding a maximum number of discrete steps, denoted as $N_{\text{max}}$.
This stop action signals that the agent has recognized the successful completion of the navigation task.

\begin{figure}
    \centering
    \includegraphics[width=0.8\linewidth]{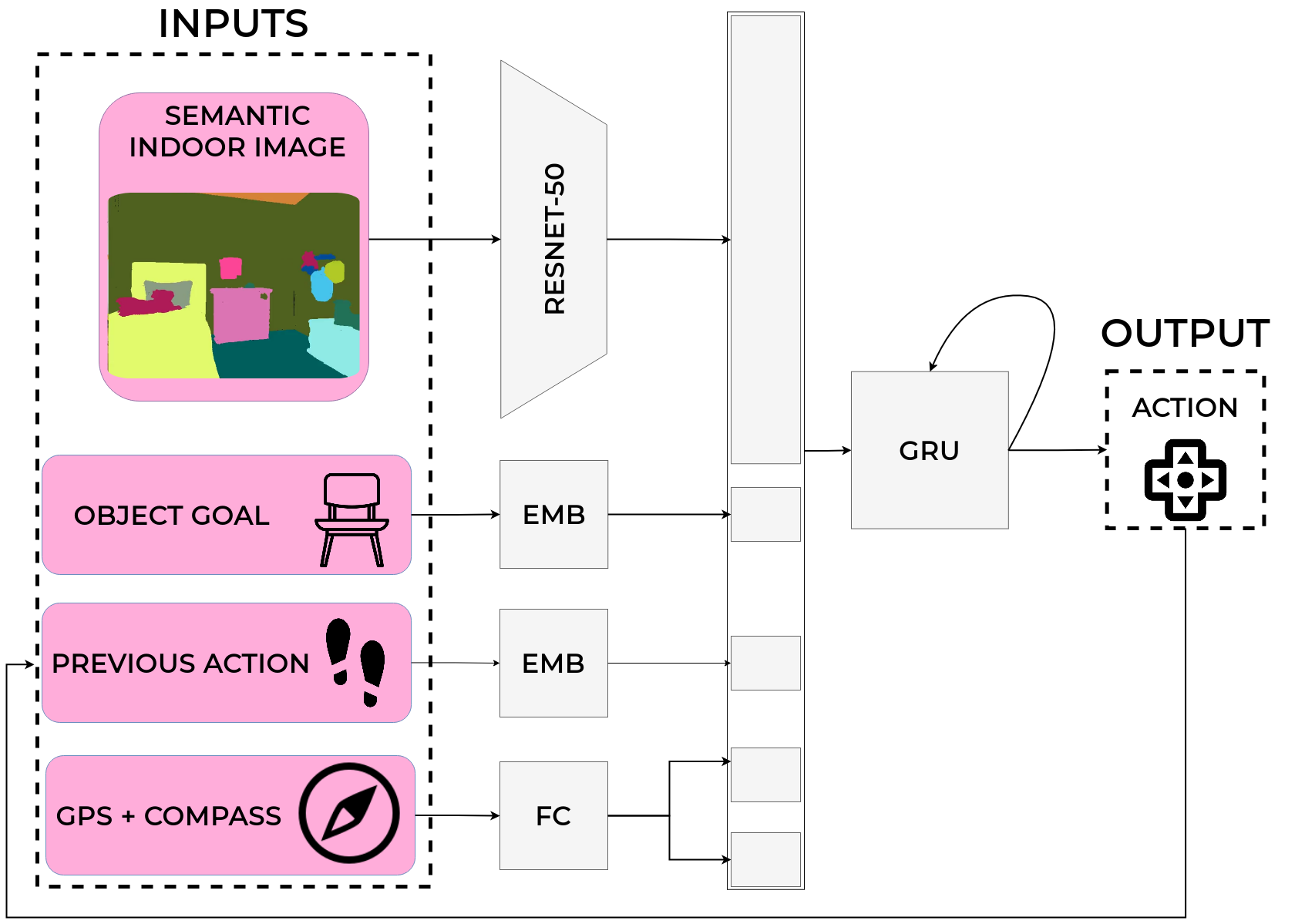}
    \caption{Our proposed architecture for the \semnav model, where the main visual input is a semantic segmentation.}
    \label{fig:scheme_semnav}
\end{figure}

For our \semnav model, we define the navigation task as follows.
Given a target object class $c_i$, e.g., \emph{chair}, our goal is to navigate to an instance of this category using
only the semantic segmentation of the environment, within a  maximum of 500 discrete actions $N_{\text{max}}=500$.
Figure~\ref{fig:scheme_semnav} shows the architecture of the proposed \semnav model, which leverages semantic segmentation information as input to guide the navigation policy.
We define a VSN model in which, at each position $p_i$, the agent has access to an environmental observation $o_{p_i}$, represented as the tuple:
\[
o_{p_i} = \{o^{\text{semantic}}_i, o^{\text{ori}}_i, o^{\text{pos}}_i\} \;.
\]
    
As shown in Figure \ref{fig:scheme_semnav}, the \semnav model receives as inputs: the semantic segmentation of the environment \(o^{\text{semantic}}_i\); the relative orientation with respect to the initial viewpoint \(o^{\text{ori}}_i = \Delta \alpha_i\); the relative displacement from the starting position \(o^{\text{pos}}_i = (\Delta x_i, \Delta y_i, \Delta z_i)\); the previous action taken by the agent \(a_{i-1}\); and the target object class \(c_i\).

\[
o_{p_i} = \{\,o^{\text{semantic}}_i,\; o^{\text{ori}}_i,\; o^{\text{pos}}_i,\; a_{i-1},\; c_i\,\}.
\]

The output of our \semnav model is a discrete navigation action that moves the agent within the environment.
Formally, we define our \semnav model as a mapping between an observation $o_{p_i}$ and an action $a_{p_i}$ within the discrete set $\mathcal{A} = \{ \turnleft, \turnright, \moveforward, \allowbreak \movebackward, \stopac\}$.
This mapping is learned through a navigation policy implemented by a deep learning model, as depicted in Figure \ref{fig:scheme_semnav}.
Let $\pi^{\semnav}_{\theta}(a_{p_i} \mid o_{p_i})$ denote this navigation policy, parameterized by the weights $\theta$ of the network architecture responsible for mapping observations to a probability distribution over actions in $\mathcal{A}$.

For the \semnav model, we adopt an IL strategy based on human demonstrations to learn the parameters in $\theta$.
Specifically, in our experiments, we leverage a dataset of 77k human-driven navigation trajectories from HM3D~\cite{ramrakhya2022}, which are also applicable to the novel \semnav dataset.
A trajectory $\tau$ of length $T$, representing a human demonstration, is defined as a sequence of observation-action tuples:$\tau = (o_{p_0}, a_{p_0}, o_{p_1}, a_{p_1}, \dots, o_{p_T}, a_{p_T})$.
The set of trajectories used for training is denoted as $\mathcal{T} = \big\{\tau_i\big\}_{i=1}^N$.
Following an IL approach, we optimize the parameters $\theta^*$ of our navigation policy by solving the following optimization problem
\begin{equation}
    \theta^* = \argmin_{\theta} \sum_{i=1}^{N} \sum_{(o_{p_t},a_{p_t}) \in \tau_i}^{T} -\log(\pi^{\semnav}_{\theta}(a_{p_t} \mid o_{p_t})) \;,
\end{equation}
where the navigation policy is trained by minimizing the discrepancy between the probability distribution for the actions assigned by the model to each observation $o_{p_t}$, and the actions provided by the human experts.

The architecture depicted in Figure \ref{fig:scheme_semnav} comprises several key components.
To process the semantic segmentation input, we first transform it into an RGB format.
Each pixel is assigned a semantic segmentation label, which is then mapped to a specific color in the RGB space using the semantic segmentation sensor integrated into the \semnav dataset.
The encoded input is subsequently processed by a ResNet-50 architecture~\cite{resnet}, which extracts high-level visual features for downstream tasks. This process is formalized in the following equation:
\begin{equation}
h^{res}_i = f_{\text{RESNET}}(o^{\text{semantic}}_i; \theta_{\text{RESNET}}) \in \mathbb{R}^{d_{\text{res}}},
\end{equation}
where $h^{\text{res}}_i$ represents the extracted visual feature vector at position $p_i$, $f_{\text{RESNET}}(\cdot)$
denotes the ResNet-50 feature extraction function, $o^{\text{semantic}}_i$ is the semantic segmentation input at position $p_i$, and $\theta_{\text{RESNET}}$ encodes the learnable parameters of the ResNet-50. Here, the output dimensionality of the ResNet-50 is $d_{\text{res}} = 512$, meaning that the resulting feature vector belongs to a 512-dimensional real-valued space.

For the experimental evaluation, we also used architectures that incorporate multiple visual modalities.
Technically, we are able to integrate both semantic segmentation and RGB images as inputs.
In this case two parallel ResNet-50 backbones are employed, each dedicated to processing one modality independently.
This architectural design was selected not only for its favorable balance between computational efficiency and performance, but also to ensure a fair and consistent comparison with other ObjectNav models that adopt the same configuration, thereby enabling a more equitable evaluation framework.

In our experiments, we evaluate two types of initialization for the ResNet50: (1) a random initialization and (2) an initialization using DINO~\cite{Caron2021EmergingPI}.

The input corresponding to the target object category is represented by its class index $c_i$ and processed through an embedding layer, as shown in the following equation:
\begin{equation}
h^{\text{goal}}_i = W_{\text{goal}}[c_i] \in \mathbb{R}^{d_{\text{goal}}},
\end{equation}
where $h^{\text{goal}}_i$ denotes the embedded representation of the target category at timestep $i$, and
$W_{\text{goal}} \in \mathbb{R}^{m \times d}$ represents the learnable embedding matrix, where $m = 6$ is the number of object categories and $d_{\text{goal}} = 32$ is the embedding dimension. The bracket notation $[\cdot]$ indicates the selection of the corresponding row from the embedding matrix associated with category $c_i$.

The position and orientation observations, both expressed relative to the starting position, are processed through independent fully connected layers. The position observation is projected from $\mathbb{R}^2$ to a $d_{\text{pos}}$-dimensional embedding space, and the orientation observation, represented by the cosine and sine of the relative heading angle, is projected from $\mathbb{R}^2$ to a $d_{\text{ori}}$-dimensional embedding space. This process is formalized in the following equations:
\begin{equation}
h^{pos}_i = W_{pos} \, o^{pos}_i + b_{pos} \in \mathbb{R}^{d_{\text{pos}}},
\quad
h^{ori}_i = W_{ori} \, o^{ori}_i + b_{ori} \in \mathbb{R}^{d_{\text{ori}}}\; .
\end{equation}

Here, $o^{pos}_i \in \mathbb{R}^2$ and $o^{ori}_i \in \mathbb{R}^2$ denote the position and orientation observations at timestep $i$, respectively. The learnable parameters $W_{pos}, W_{ori} \in \mathbb{R}^{d_{\text{pos}}\times 2}, \mathbb{R}^{d_{\text{ori}}\times 2}$ and $b_{pos}, b_{ori} \in \mathbb{R}^{d_{\text{pos}}}, \mathbb{R}^{d_{\text{ori}}}$ correspond to the weights and biases of the linear embedding layers. In our implementation, $d_{\text{pos}} = d_{\text{ori}} = 32$, and the resulting feature vectors $h^{pos}_i, h^{ori}_i$ are then concatenated with the other observation embeddings before being passed to the recurrent module.

To additionaly incorporate the agents previous action into the policy input, an embedding of the action taken at the previous timestep is included. Let $a_{i-1}$ denote the previous action and $W_{act}\mathbb{R}^ {n_{act} \times d_{act}}$ be the learnable embedding matrix where $n_{act} = 6$ is the number of discrete actions and $d_{act}$ is the action embedding dimensionality.
The previous action embedding is obtained as:
\begin{equation}
h^{\text{act}}_{i} = W{\text{act}}[a_{i-1}] \in \mathbb{R}^{d_{\text{act}}}.
\end{equation}
All the processed observations, including the visual embedding from the ResNet-50, the object goal embedding, and the embeddings from the position and orientation, are concatenated into a single feature vector at each timestep:
\begin{equation}
x_i = \big[ h^{res}_i \,;\, h^{emb}_i \,;\, h^{pos}_i \,;\, h^{ori}_i \,;\, h^{act}_i \big] \in \mathbb{R}^{d_x},
\end{equation}
where $d_x = d_{res} + d_{emb} + d_{pos} + d_{ori} + d_{act}$ denotes the total dimensionality of the concatenated input, and $[\,;\,]$
represents vector concatenation.

This vector $x_i$ is then passed through a Gated Recurrent Unit (GRU)~\cite{cho-etal-2014-learning} to capture temporal
dependencies and provide the model with a memory of past navigation decisions. In our implementation, the GRU has
two recurrent layers ($\text{num\_recurrent\_layers}=2$) and a hidden size of $d_{h} = 2048$, as specified in the
configuration:
\begin{equation}
h_{\text{i}} = \text{GRU}(x_i, h_{\text{i}-1}; \theta_{\text{i}}) \in \mathbb{R}^{d_{h}},
\end{equation}
where $h_{i} \in \mathbb{R}^{d_{h}}$ is the hidden state of the GRU at timestep $i$,
$d_{h} = 2048$ denotes the dimensionality of the hidden layer, and $\theta_{\text{GRU}}$ represents the learnable
parameters of the GRU across both layers. The GRU allows the model to condition its current action predictions on past observations without the need for an explicit map, enabling autonomous, map-free navigation.
This way, the recurrent hidden state $h_i$ conditions the model's action predictions, by taking into account the history of past observations and navigation actions. The GRU output is subsequently used to predict the discrete action distribution that forms the output of our policy.

The hidden state $h_i \in \mathbb{R}^{d_h}$ produced by the GRU at timestep $i$ is subsequently fed into a
multi-layer perceptron (MLP) to produce the unnormalized action logits. Let $n_a$ denote the number of discrete actions; in our case, $n_a = 6$, because our action space is $\mathcal{A} = \{ \turnleft, \turnright, \moveforward, \allowbreak \movebackward, \stopac\}$. The process can be formalized as follows:
\begin{equation}
z_i = f_{\text{MLP}}(h_i; \theta_{\text{MLP}}) \in \mathbb{R}^{n_a},
\end{equation}
where $f_{\text{MLP}}(\cdot)$ represents the multi-layer perceptron with learnable parameters $\theta_{\text{MLP}}$, and $z_i$ are the logits corresponding to each discrete action.

These logits are then passed through a categorical distribution to obtain the action probabilities:
\begin{equation}
\pi(a_{p_i} \mid o_{p_i}) = \text{Categorical}(\text{softmax}(z_i)) \in \mathbb{R}^{n_a},
\end{equation}
where $\pi(a_{p_i} \mid o_{p_i})$ defines the probability distribution over the $n_a$ possible actions given all
observations up to timestep $i$, and the \text{softmax} function ensures a valid probability distribution.

In summary, the recurrent hidden state $h_i$ encodes the temporal context and is transformed by the MLP to produce logits, which are subsequently normalized by the softmax function to define a categorical policy over the discrete action space.

The complete training procedure for the \semnav visual navigation policy is summarized in Algorithm~\ref{alg:semnav_training}, which formalizes the imitation learning process described above.
The algorithm details the iterative optimization of the network parameters $\theta$ using expert demonstrations from the dataset $\mathcal{T}$, encompassing the visual encoding, feature embedding, recurrent processing, and action prediction steps at each timestep.
This end-to-end learning framework enables the agent to acquire robust navigation policies that leverage semantic segmentation information to navigate efficiently toward target object categories in previously unseen environments.

\begin{algorithm}[t]
\caption{Imitation Learning for \semnav Visual Navigation Policy}
\label{alg:semnav_training}
\begin{algorithmic}[1]
{\footnotesize
\REQUIRE $\mathcal{T} = \{\tau_i\}_{i=1}^N$: set of expert demonstrations
\REQUIRE
    $\theta = \{\theta_{\text{RESNET}}, \theta_{\text{GRU}}, \theta_{\text{MLP}}, W_{\text{goal}}, W_{\text{pos}}, W_{\text{ori}}, W_{a_{i-1}\}}$: model parameters
\REQUIRE $\alpha$: learning rate, $B$: batch size, $E$: number of epochs
\STATE Initialize $\theta$ randomly or with DINO pre-training for $\theta_{\text{RESNET}}$
\FOR{epoch $e = 1$ to $E$}
    \STATE Shuffle trajectories $\mathcal{T}$
    \FOR{each mini-batch of $B$ trajectories}
        \STATE Initialize gradient accumulator $\nabla_\theta \mathcal{L} \leftarrow 0$
        \FORALL{trajectory $\tau = (o_{p_0}, a_{p_0}, \dots, o_{p_T}, a_{p_T}) \in \text{batch}$}
            \STATE Initialize recurrent state $h_0 \in \mathbb{R}^{d_h}$
            \FORALL{timestep $t = 0$ to $T$}
                \STATE \textbf{// Visual encoding}
                \STATE $h^{\text{res}}_t \leftarrow f_{\text{RESNET}}(o^{\text{semantic}}_t; \theta_{\text{RESNET}})$
                \STATE \textbf{// Goal, position, previous action and orientation embeddings}
                \STATE $h^{\text{goal}}_t \leftarrow W_{\text{goal}}[c_i]$
                \STATE $h^{\text{pos}}_t \leftarrow W_{\text{pos}} \, o^{\text{pos}}_t + b_{\text{pos}}$
                \STATE $h^{\text{ori}}_t \leftarrow W_{\text{ori}} \, o^{\text{ori}}_t + b_{\text{ori}}$
                \STATE $h^{\text{act}}_t \leftarrow W{\text{act}}[a_{i-1}]$
                \STATE \textbf{// Concatenate features}
                \STATE $x_t \leftarrow [h^{\text{res}}_t; h^{\text{goal}}_t; h^{\text{pos}}_t; h^{\text{ori}}_t]$
                \STATE \textbf{// Recurrent processing}
                \STATE $h_t \leftarrow \text{GRU}(x_t, h_{t-1}; \theta_{\text{GRU}})$
                \STATE \textbf{// Action prediction}
                \STATE $z_t \leftarrow f_{\text{MLP}}(h_t; \theta_{\text{MLP}})$
                \STATE $\pi_\theta(a_{p_t} \mid o_{p_t}) \leftarrow \text{Categorical}(\text{softmax}(z_t))$
                \STATE \textbf{// Accumulate loss}
                \STATE $\nabla_\theta \mathcal{L} \leftarrow \nabla_\theta \mathcal{L} - \nabla_\theta \log \pi_\theta(a_{p_t} \mid o_{p_t})$
            \ENDFOR
        \ENDFOR
        \STATE \textbf{// Update parameters}
        \STATE $\theta \leftarrow \theta - \alpha \cdot \nabla_\theta \mathcal{L} / B$
    \ENDFOR
\ENDFOR
\RETURN Trained policy parameters $\theta$
}
\end{algorithmic}
\end{algorithm}

\section{Experiments}
\label{sec:experiments}

\subsection{Experimental settings and evaluation metrics}

Our main goal in this work is to develop VSN solutions capable of navigating efficiently both in the virtual environments \emph{and} in the real world.
For this purpose, we have designed two types of experiments.

\paragraph{Experiments in the \textbf{simulation environment}}
For this experimental evaluation, we addressed the \objnav task detailed in the Habitat 2023 challenge~\cite{habitatchallenge2023}.
We used the official HM3D~\cite{yadav2022habitat} training and validation sets.
For our \semnav model we employed the two semantic segmentation sensors detailed in Section \ref{sec:dataset_semnav}.
%We used the 77k expert annotations available in \cite{Ramakrishnan2021HabitatMatterport3D} for training our \semnav models.
For evaluation metrics, we followed the standard proposals for the \objnav problem~\cite{habitatchallenge2023}.
Specifically, these metrics include: 1) the Success Rate (SR), defined as the percentage of navigation episodes successfully completed by the model; 2) the Shortest Path Length (SPL), which compares the distance traveled by the agent during the episode with the shortest possible path length to the target object; 3) the average number of collisions the agents experience with the environment (C); and 4) the average distance to the goal (DTG), which is the mean distance of the agent from the nearest target object at the end of a navigation episode.
A navigation episode is considered successful when the agent samples the \stopac action within 1 meter of the target object it is tasked to find.

\paragraph{Experiments in the \textbf{real world}}

\begin{figure}
    \centering
    \subfloat[House 1]{\includegraphics[width=0.32\linewidth]{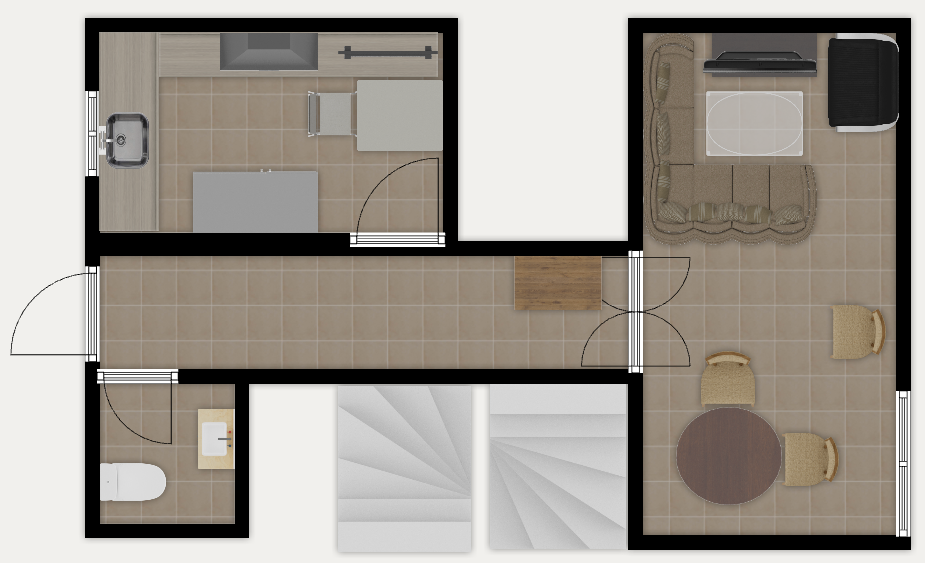}\label{fig:house1}}\hfill
    \subfloat[House 2]{\includegraphics[width=0.32\linewidth]{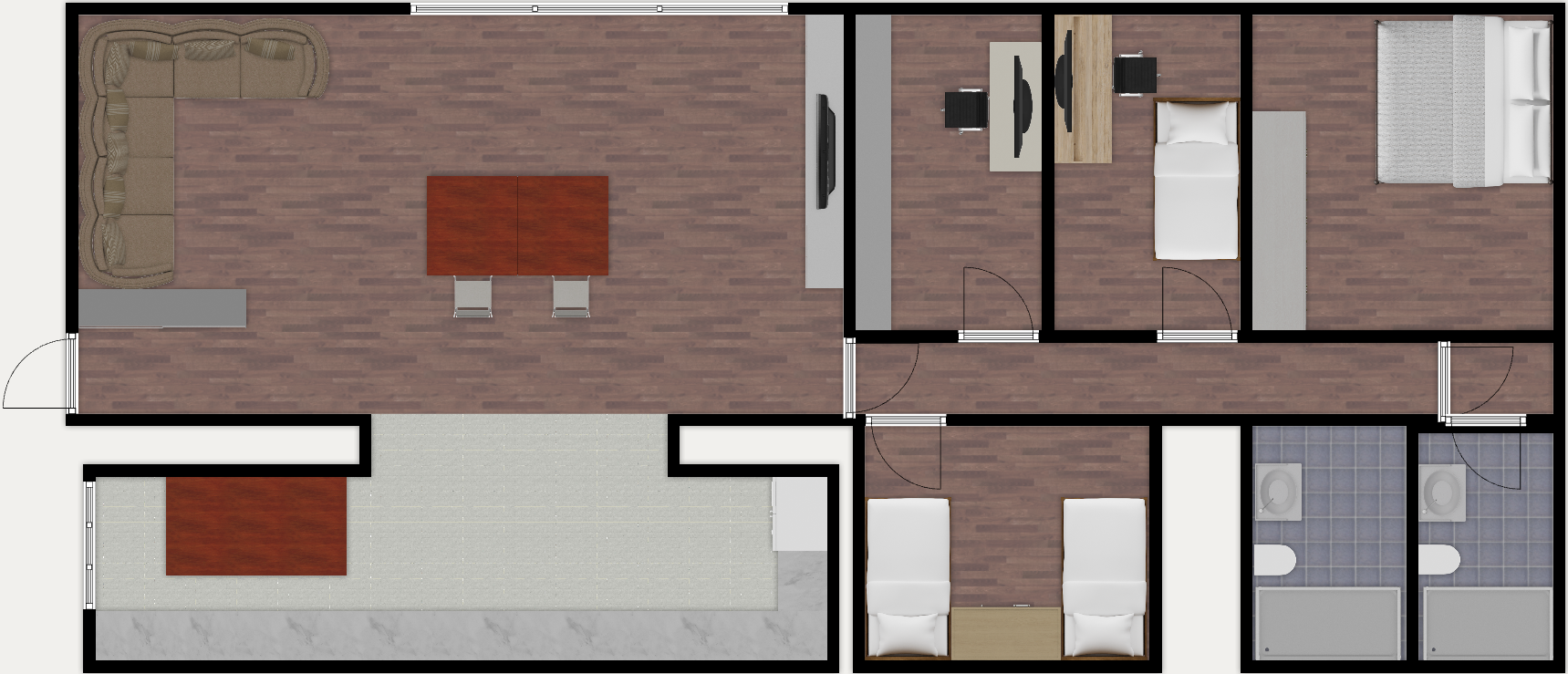}\label{fig:house2}}\hfill
    \subfloat[House 3]{\includegraphics[width=0.32\linewidth]{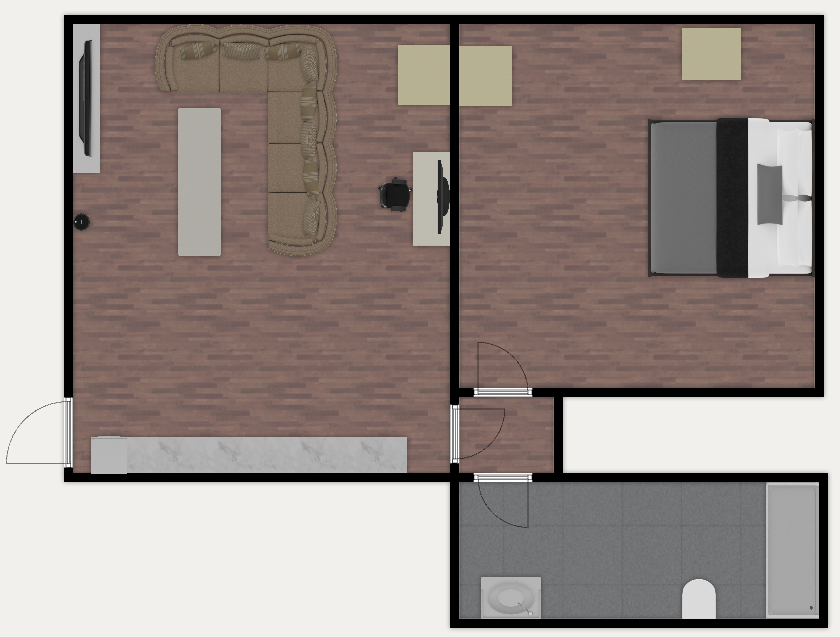}\label{fig:house3}}
    \caption{Top-down views of the houses where \objnav real-world experiments were conducted for five object categories.}
    \label{fig:houseforexperiments}
\end{figure}

To assess the performance of our models in the real world, experiments were conducted in three different houses, whose floor plans are shown in Figure~\ref{fig:houseforexperiments}.
That is, the \semnav model, trained entirely in simulation, is now evaluated in the real world—within three unseen environments that present a significant domain gap.
This type of experiment is essential to assess whether our proposed \semnav approach can generalize effectively and navigate reliably in real-world conditions.

We employed a TurtleBot 2 robotic platform~\cite{kobuki} specifically adapted for the \objnav problem.
This adaptation replicates the agent's characteristics during training.
Among the modifications, we added a mast to the TurtleBot 2, raising the camera to 1.25 m to match the simulation setup.
The camera used is an Orbbec Astra with depth perception.
Since our \semnav model requires a semantic segmentation sensor, for real-world testing, we integrated the ESANet model~\cite{Seichter2020EfficientRS}, pre-trained on the NYUv2 dataset~\cite{Silberman2012IndoorSA}.
This model takes the robot's egocentric vision as input in RGB+Depth format and generates a semantic segmentation image, which is then fed into our \semnav model.
Each segmentation inference incurs an additional overhead of approximately 11ms per frame when running on an NVIDIA RTX4080.
In these real-world experiments, we start with a detailed comparative study of performance between our proposed \semnav model and state-of-the-art VSN approaches, incorporating the PirlNav model~\cite{ramrakhya2023} into the evaluation.
This comparative experiment was carried out in the house 1, whose floor plan can be seen in Figure~\ref{fig:house1}.
We analyze in house 1, the main types of errores that our models exhibit, as well as for the state-of-the-art PirlNav approach.
Then, we extend the whole experimental comparison to houses 2 and 3, see Figures~\ref{fig:house2} and~\ref{fig:house3}, respectively.
Our best-performing model in simulation, \semnav-RGBS+DINO $\rightarrow$ RL, was evaluated in all these houses.

The embedding of our \semnav model and the state-of-the-art PirlNav~\cite{ramrakhya2023} in the robotic platform was done using the ROS4VSN library~\cite{GutierrezAlvarez2023VisualSN}, which allows for the integration of VSN solutions into real robots using ROS.
%We also followed the real-world testing protocol described in~\cite{GutierrezAlvarez2023VisualSN}.
We conducted navigation experiments toward five specific object classes: chair, bed, toilet, television monitor, and sofa.
For each episode, an object category was selected as the navigation target, and the navigation started from a predefined location and orientation.
These starting points were chosen to ensure that: 1) there was no direct visibility of the target object; and 2) there was a minimum distance of 10 meters to the object.
The evaluation metric used was the SR, where a navigation episode is considered successful if the robot infers the stop action while being within one meter of the target object.
A failure is recorded if the robot collides and cannot proceed or if it requires more than 225 actions to complete the task.
In the real world, $N_{\text{max}}=225$, as real environments typically offer less navigable surface compared to simulation.

Calculating the SPL metric in the real world is challenging, so we introduce a new metric, Success Divided by Step (SDS), as an alternative for evaluating navigation efficiency in real environments.
SDS is computed as $SDS = \frac{N_{TS}}{N_{AS}}$, where $N_{TS}$ is the number of successful trajectories, and $N_{AS}$ is the total actions taken during those successful trajectories to reach the target object category.

We release all code for training and evaluating our models, both in simulated environments and in the real world, in the \href{https://github.com/gramuah/semnav}{following repository}.

\subsection{Results in simulation}

\subsubsection{Ablation study}
In this analysis, we aim to present the impact of the following aspects on the performance of our \semnav model: 1) We test a \semnav configuration that uses only semantic segmentation as visual input—\semnav-Only Semantic Segmentation (\semnav-OS); 2) We add a standard RGB visual input to the \semnav model, which passes through its corresponding visual encoder (a ResNet 50) to be concatenated with the rest of input features—\semnav-RGB + Semantic Segmentation (\semnav-RGBS); 3) Each of these configurations is tested with semantic segmentation using 1630 and 40 categories, as provided in the two \semnav dataset sensors; and 4) For each test, we train the model either from randomly initialized weights or by pre-initializing them using DINO~\cite{Caron2021EmergingPI}, since this initialization strategy has demonstrated a significant impact on previous VSN models, e.g., PirlNav~\cite{ramrakhya2023}.
In total, we propose an evaluation of up to eight different configurations for our \semnav model.
\begin{table}[t]
    \centering
    \resizebox{\linewidth}{!}{%
    \begin{tabular}{ll|cccc}
        Categories dataset & Model & \textbf{SR (\textuparrow)} & \textbf{SPL (\textuparrow)} & \textbf{C (\textdownarrow)} & \textbf{DTG (\textdownarrow)} \\
        \toprule
        \multirow{5}{*}{\semnav1630}    & PirlNav (RGB) & 60.9 & 0.26 & 45.74 & 3.17 \\
                                        & \semnav-OS & 65.1 & 0.29 & 57.11 & 2.99\\
                                        & \semnav-OS+DINO & 68.4 & 0.30 & 50.39 & 2.75\\
                                        & \semnav-RGBS & 69.2 & 0.31 & 50.59 & 2.73\\
                                        & \semnav-RGBS+DINO & \cellcolor{pink!60}70 & \cellcolor{pink!60}0.32 & \cellcolor{pink!60}39.17 & \cellcolor{pink!60}2.69\\

        \midrule
        \multirow{5}{*}{\semnav40}      & PirlNav (RGB) & 60.9 & 0.26 & 45.74 & 3.17 \\
                                        & \semnav-OS & 72.9 & 0.34 & 43.25 & 2.55\\
                                        & \semnav-OS+DINO & 73 & 0.34 & 48.49 & 2.61\\
                                        & \semnav-RGBS & 74.3 & 0.35 & 40.70& 2.50\\
                                        &  \semnav-RGBS+DINO & \cellcolor{pink!60}76.2 & \cellcolor{pink!60}0.36 & \cellcolor{pink!60}38.50 & \cellcolor{pink!60}2.36 \\

        \bottomrule
    \end{tabular}
    }
    \caption{Evaluation of the performance of different configurations of our \semnav model. We report the metrics: SR, SPL, C, and DTG.}
    \label{table:ablation_study}
\end{table}

%Discussion
Table~\ref{table:ablation_study} presents the performance of all these configurations on the \semnav-dataset validation set.
In light of the results obtained in this study, our model \semnav-RGBS has provided the best
performance.
It is interesting to highlight that all models trained with 40-category semantic segmentations have outperformed those trained with 1630-category semantic segmentations.
We believe this is primarily due to the annotation issues present in the 1630-category set, as detailed in Section \ref{sec:dataset_semnav}, and the fact that the semantic information necessary for successful navigation is already included in the 40 categories, making the 1630-category segmentation unnecessarily complex.
Lastly, our \semnav model does not appear to be significantly affected by initialization with DINO~\cite{Caron2021EmergingPI}, unlike other state-of-the-art models.
In our case, DINO initialization slightly improves the results.

This ablation study also demonstrates that semantic segmentation is the key factor impacting model performance.
Comparing any of our models with and without RGB input—i.e., \semnav-OS vs. \semnav-RGBS—further
confirms
this observation.
For instance, in terms of SR, for the \semnav40 dataset, the \semnav model improves from 72.9 to 74.3 when using the \semnav-OS and \semnav-RGBS versions, respectively.
We conclude that these results validate the suitability of the \semnav approach for the VSN problem.
Adding RGB information to the model consistently enhances its performance, and for real-world navigation, this would be the optimal architecture to embed in a robotic platform, as we will demonstrate in Section \ref{sec:real_world_results}.

\subsubsection{Semantic segmentation precision impact}

The training of the \semnav models has been conducted using the semantic segmentation information provided by our novel semantic sensors in the \semnav dataset.
These semantic segmentations are derived from manually annotated data, which provides high accuracy.
However, we consider it important to evaluate the impact that the quality of semantic segmentation may have on navigation performance.
This section presents a study on the impact of using semantic segmentations of varying quality through the following experiments.
Technically, we propose to tweak the original semantic segmentation quality in the \semnav sensor 40 by injecting three types of noise to mimic imperfections that could appear in a real-world application:
\begin{enumerate}
    \item \textbf{Category misclassification noise}:
    This noise model introduces semantic confusion by injecting probabilistic label errors.
    Each semantic class is assigned a probability of being misclassified as another class.
    For example, pixels belonging to the category chair may be incorrectly labeled as table or floor, emulating the typical misclassification patterns observed in automatic segmentation systems when differentiating visually similar objects.
    \item \textbf{Boundary noise}: This perturbation degrades boundary accuracy by introducing spatial and temporal distortions along segmentation edges.
    Unlike the sharp, well-defined contours in ground-truth annotations, this noise generates irregular, jagged boundaries that fluctuate across consecutive frames, effectively modeling the uncertainty and temporal inconsistency observed in real-world semantic segmentation systems, particularly near object transitions and occlusion regions.
    \item \textbf{Spatially localized perturbations}: This noise introduces discrete, spatially confined artifacts by randomly sampling patches of incorrect labels across the image.
    These perturbations emulate localized segmentation failures, such as misclassified regions, spurious detections,
    or missing object parts.
    These are typical artifacts in real-world semantic segmentation systems caused by challenging illumination, texture ambiguities, or sensor-induced noise.
\end{enumerate}

In this study on the influence of semantic segmentation accuracy, we focus primarily on \textbf{category misclassification noise}, as it represents the most prevalent error type in both simulated (arising from dataset misannotations) and real-world semantic segmentation systems.
We prioritize this noise type in our analysis before extending the evaluation to boundary and spatially localized perturbations.

\begin{table}[t]
    \centering
    \resizebox{0.5\linewidth}{!}{%
    \begin{tabular}{l|cccc}
        \textbf{SR with sensor errors} & \textbf{0\%} & \textbf{1\%} & \textbf{10\%} & \textbf{30\%} \\
        \toprule
        \textbf{SR} & 76.2 &  74.5 &  73.3& 69.9   \\
        \textbf{SPL} & 0.36 &  0.353 &  0.337& 0.287   \\
        \bottomrule
    \end{tabular}
    }
    \caption{Evaluation of SR and SPL under different levels of Category misclassification sensor noise.}
    \label{table:sr_sensor_errors}
\end{table}

\begin{figure}[h]
  \centering
  \setlength{\tabcolsep}{4pt} % reduce horizontal gap between columns
  \renewcommand{\arraystretch}{0.98} % reduce vertical gap between rows
   \resizebox{\linewidth}{!}{%
  \begin{tabular}{ccc}
    \subfloat[]{\includegraphics[width=0.325\textwidth]{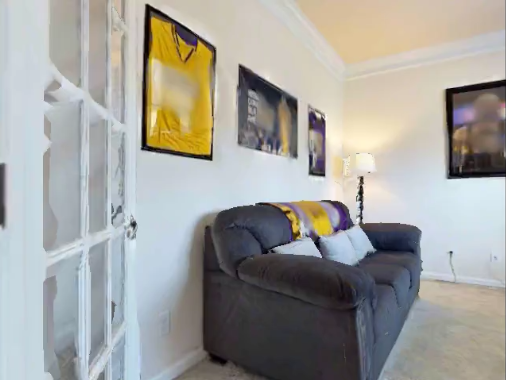}} &
    \subfloat[]{\includegraphics[width=0.325\textwidth]{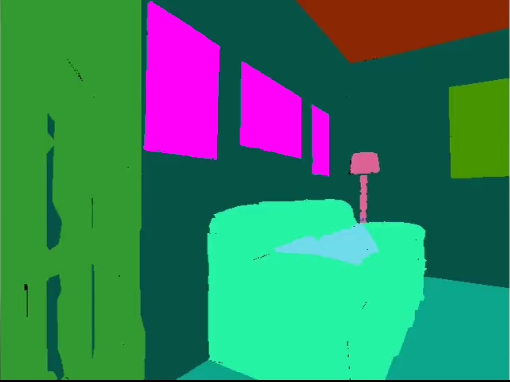}\label{fig:impact_b}} &
    \subfloat[]{\includegraphics[width=0.325\textwidth]{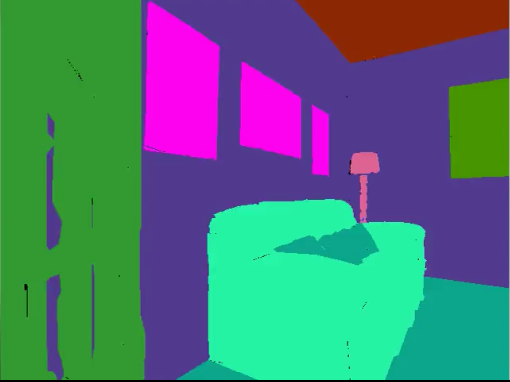} \label{fig:impact_c}}\\[-2pt]
    \subfloat[]{\includegraphics[width=0.325\textwidth]{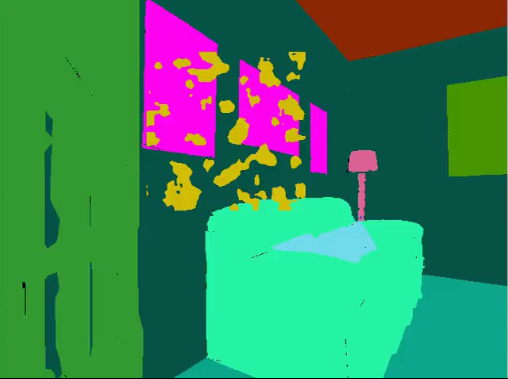}} &
    \subfloat[]{\includegraphics[width=0.325\textwidth]{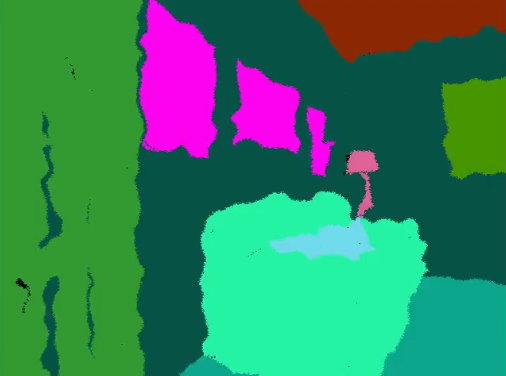}}&
    \subfloat[]{\includegraphics[width=0.325\textwidth]{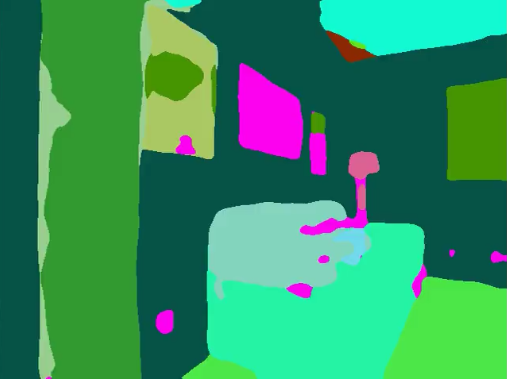}}\\[-2pt]
    \multicolumn{3}{c}{\subfloat[]{\includegraphics[width=0.325\textwidth]{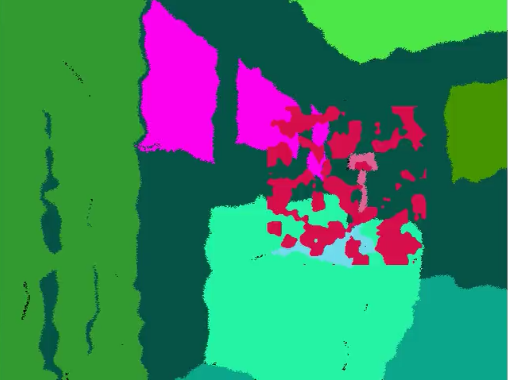}}}
  \end{tabular}
  }
  \caption{Qualitative results in the simulation environment for the different noisy semantic segmentation sensors. (a) RGB sensor output overlaid on the ground truth, (b) semantic segmentation sensor with no noise, (c) semantic segmentation missclassification noise with 10\% error, (d) spatially localized perturbations noise error, (e) Border Error noise, (f) ESANet output, (g) semantic segmentation missclassification noise, spatially localized perturbation noise error and border error noise combined.}
  \label{fig:semantic_impact}
\end{figure}

Table~\ref{table:sr_sensor_errors} shows the impact that this type of noise has on the metrics used to evaluate navigation performance.
Technically, in this experiment we have used noisy semantic segmentation sensors with varying misclassification probabilities of $0\%$, $1\%$, $10\%$, and $30\%$.
The noise injection process operates as follows: when a batch of images is received, each semantic category within the batch has a certain probability of being incorrectly relabeled as a different category from the 40 available classes.
Critically, when a category is selected for misclassification, \emph{all pixels belonging to that category} are simultaneously relabeled to the same alternative category.
This probabilistic misclassification is applied at the batch level, meaning the category substitutions vary across different batches and do not persist throughout the entire navigation episode.
Consequently, the spatial structure and boundaries of the segmentation remain intact, as they are directly derived from the ground truth annotations; only the semantic labels are altered coherently across all pixels of each affected category.
Figure~\ref{fig:impact_c} illustrates the visual effect of category misclassification noise at a $10\%$ error rate.
As shown, the wall pixels have been incorrectly relabeled, resulting in a different color assignment compared to the ground-truth semantic segmentation depicted in Figure~\ref{fig:impact_b}.

Figure~\ref{fig:semantic_impact} (c) illustrates the visual effect of this category misclassification noise.

For the specific evaluation of the impact of the noisy semantic segmentations, the performance was assessed using SR, SPL, C, and DTG metrics with our \semnav-RGBS+DINO model trained via IL.
These metrics are reported using 2000 navigation trajectories.
The results of this experiment are presented in Table~\ref{table:sr_sensor_errors}.
As the misclassification rate of the semantic segmentation sensor increases, both SR and SPL metrics degrade, highlighting the sensitivity of navigation performance to segmentation quality.
This experiment yields two main conclusions.
First, it underscores the critical role of semantic information in navigation tasks: higher semantic segmentation accuracy directly correlates with improved navigation performance.
Second, when deploying models such as \semnav in real-world scenarios—where accurate or manually annotated semantic segmentations are unavailable—a degradation in navigation performance might be expected.

What happens when the remaining types of noise are taken into consideration?
In this second experiment, each noise type was independently injected into the semantic segmentation sensor, as well as a combined configuration including all noise types.
Results over 2000 evaluation trajectories are summarized in Table~\ref{tab:allerrors}.

\begin{table}[t]
    \centering
    \resizebox{0.7\linewidth}{!}{%
    \begin{tabular}{l|cccc}
        \textbf{Noise Type} & \textbf{SR (\%)} & \textbf{SPL} & \textbf{C} & \textbf{DTG} \\
        \toprule
        No Noise & 76.2 & 0.36 & 38.50 & 2.36 \\
        \midrule
        Boundary noise & 74.5 & 0.352 & 38.6 & 2.47 \\
        10\% Category misclassification & 73.3 & 0.337 & 40.73 & 2.5 \\
        Spatially localized perturbations & 68.95 & 0.288 & 44.2 & 2.79 \\
        \midrule
        All Noises Combined & 64.05 & 0.252 & 48.49 & 3.07 \\
        \bottomrule
    \end{tabular}
    }
    \caption{Evaluation of SR, SPL, C, and DTG under different types of semantic segmentation sensor noise.}
    \label{tab:allerrors}
\end{table}

As shown in Table~\ref{tab:allerrors}, introducing noise into the semantic segmentation negatively impacts the navigation performance of the \semnav model, regardless of the noise type.
Boundary segmentation errors cause a moderate decrease in performance, whereas category misclassification noise and spatially localized perturbations have a more pronounced effect.

Boundary segmentation errors cause a moderate decrease in performance, whereas category misclassification noise and spatially localized perturbations have a more pronounced effect.
This is because abrupt changes in object geometry do not fundamentally alter the semantic structure of the navigated space, whereas category misclassifications or spurious category labels can mislead the agent into reasoning that it is in a semantically different location.
For instance, if the agent observes pixels or objects labeled as sofa within a bathroom, this semantic
inconsistency significantly disrupts the navigation process.

When all noise types are combined, the model’s performance degrades substantially, highlighting the importance of accurate semantic segmentation for optimal performance in visual navigation tasks.
The qualitative impact of the proposed noise models is illustrated in Figure~\ref{fig:semantic_impact}, which shows examples of semantic segmentations under different noise configurations.
The applied perturbations generate simulated artifacts that could resemble those produced by automatic semantic segmentation models in real-world conditions.
This similarity is particularly evident in the comparison between images (f) and (g), where the noise introduced by the actual segmenter is contrasted with the combined effect of the three proposed noise types.

The segmentation in (f) produced by ESANet~\cite{Seichter2020EfficientRS} exhibits category misclassification noise, observable in the floor, ceiling, and wall paintings; boundary errors, evident in the less sharp floor edges compared to noise-free simulation; and spatially localized perturbations, manifested as spurious object categorizations within other objects, such as on the sofa or paintings.
Image (g) artificially replicates this behavior by combining all noise types, showing misclassifications (e.g., ceiling), boundary degradation across scene objects, and spatially localized perturbations in the central region.

\subsubsection{Resources analysis}
% Explanation of the computational and temporal overhead introduced by the semantic segmentation module
   
Our \semnav model needs to perform a semantic segmentation within the inference loop, therefore an additional temporal and computational
overhead is introduced during inference.
    In this experiment we perform a detailed runtime analysis, to show the impact of the use of a semantic segmentation module in our pipeline.
    Technically, as in the real-world experiments, we integrate in \semnav the ESANet~\cite{Seichter2020EfficientRS} model.
    This semantic segmentation model is utilized solely to generate the semantic segmentation image that serves as input for the \semnav-OS and \semnav-RGBS pipelines. Consequently, the total time per step and memory requirements must account for both the latency of the segmentation model and the latency of the action decision model. Table~\ref{RTXtimes} presents a temporal breakdown
per inference step, detailing the inference times for both the action models and the segmentation model. Table~\ref{tabla_memoria} summarizes the GPU memory requirements for each component (values are representative measurements obtained on an NVIDIA RTX4080).

A robotic navigation model intended for real-world deployment must exhibit low-latency temporal response, ensuring the user perceives an active system without significant delays in action execution.
For this reason, the temporal analysis presented in Table~\ref{RTXtimes} reports our model's response times.
Since the system must infer a semantic segmentation image from an RGB input, some computational overhead is introduced, which could limit practical applicability. However, as shown, although this overhead is considerably larger than the time required by the model to decide the action once all inputs are processed, the overall response times remain sufficiently low to enable more than 110 inferences per second. In practice, this inference time is negligible compared to the time required for the robot to physically execute each action, which is significantly longer than the model's inference time.

\begin{table}[ht]
    \centering
    \caption{Temporal breakdown per inference step (representative values, ms). \semnav-OS and \semnav-RGBS are action decision models; ESANet~\cite{Seichter2020EfficientRS} is the semantic segmentation model. ``Total'' = action model + semantic segmentation model.}
    \label{RTXtimes}
    \resizebox{\linewidth}{!}{%
    \begin{tabular}{lrrrr}
        \toprule
        Model & Model Inf.\ (ms) & Semantic Segmentation Inf.\ (ms) & Total per step (ms) & Approx.\ FPS \\
        \midrule
        \semnav-OS       & 1   & 7.5 & 8.5 & 117 \\
        \semnav-RGBS     & 1.5  & 7.5 & 9 & 111 \\
        \bottomrule
    \end{tabular}
    }
\end{table}

Furthermore, a memory-efficient navigation model is more suitable for deployment across diverse environments and platforms.
Table~\ref{tabla_memoria} presents the GPU memory footprint of each model evaluated in this article.
The reported sizes demonstrate that these models maintain a compact memory profile, making them compatible with a wide range of graphics hardware.

\begin{table}[ht]
    \centering
    \caption{Model size in GPU memory (MB). The reported sizes correspond to the model parameters loaded on GPU.}
    \label{tabla_memoria}
    \resizebox{\linewidth}{!}{%
    \begin{tabular}{lrr}
        \toprule
        Model & Model size on GPU (MB) & Notes \\
        \midrule
        \semnav-OS       & 188.54  & ResNet-50 backbone + head \\
        \semnav-RGBS     & 228.79  & Dual backbones (RGB + seg.) \\
        ESANet ~\cite{Seichter2020EfficientRS} & 181.27 & Semantic segmentation module \\
        \bottomrule
    \end{tabular}
    }
\end{table}

Power consumption is also a limiting factor for model deployment, particularly when integration into compact, low-power devices is desired.
Table~\ref{tab:power_consumption} reports the power consumption of navigation models on an NVIDIA RTX4080 graphics card.
As expected, training consumes significantly more power than inference, while inference power requirements remain modest and manageable for practical deployment.

\begin{table}[ht]
    \centering
    \caption{Power consumption during training and inference (representative values, W). Measurements obtained on an NVIDIA RTX4080.}
    \label{tab:power_consumption}
    \begin{tabular}{lrr}
        \toprule
        Model & Training (W) & Inference (W) \\
        \midrule
        \semnav-OS       & 115 & 13 \\
        \semnav-RGBS     & 140 & 30 \\
        \bottomrule
    \end{tabular}
\end{table}

Finally, Table~\ref{tab:bc_hyperparams} presents the hyperparameters used for training the \semnav-RGBS model via IL.
These parameters were employed on 8 NVIDIA A100 GPUs over 10 days using the \semnav 40 dataset.
These specifications highlight the computational cost of training and facilitate the reproducibility of our results.

\begin{table}[ht]
\centering
\caption{Training configuration and hyperparameters used for Behavior Cloning.}
\begin{tabular}{@{}l r@{}}
\toprule
\textbf{Parameter} & \textbf{Value} \\ \midrule
Number of GPUs & 8 \\
Number of environments per GPU & 16 \\
Rollout length & 64 \\
Number of mini-batches per epoch & 2 \\
Optimizer & Adam \\
Learning rate scheduler & Cyclic LR (exp\_range) \\
Base learning rate & $1\times10^{-5}$ \\
Maximum learning rate & $1\times10^{-3}$ \\
Step size up & 2000 \\
Exponential decay factor ($\gamma$) & 0.99994 \\
DDPIL sync fraction & 0.6 \\ \bottomrule
\end{tabular}
\label{tab:bc_hyperparams}
\end{table}

\subsubsection{Comparison with the state of the art}

We compare here the performance of our \semnav models with state-of-the-art approaches for the \objnav problem.
We have used the official validation set of HM3D~\cite{yadav2022habitat} dataset.
All compared methods use the same experimental evaluation protocol for \objnav~\cite{habitatchallenge2023}.
The only difference in terms of training data is that our \semnav models use the semantic segmentation information provided in the \semnav dataset.

\begin{table}[b]
    \centering
    \resizebox{\linewidth}{!}{%
    \begin{tabular}{l|ccc|cc}
        Model & \textbf{Depth} & \textbf{RGB} & \textbf{Semantic} & \textbf{SR (\(\uparrow\))} & \textbf{SPL (\(\uparrow\))} \\
        \toprule
        DD-PPO~\cite{wijmans2020} & $\checkmark$ & $\checkmark$ & $\times$ & 27.9 & 0.14\\
        OVRL-v2~\cite{yadav2023ovrl} & $\times$ & $\checkmark$ & $\times$ & 64.7 & 0.28\\
        Frontier based exploration~\cite{homerobot,gervet2022} & $\checkmark$ & $\checkmark$ & $\times$ & 26.0 & 0.15 \\
        Habitat-Web~\cite{ramrakhya2022} & $\checkmark$ & $\checkmark$ & $\times$ & 57.6 & 0.238\\
        PirlNav (only IL)~\cite{ramrakhya2023} (our impl.) & $\checkmark$ & $\checkmark$ & $\times$ & 60.9 & 0.26\\
        PirlNav (only IL)~\cite{ramrakhya2023} & $\times$ & $\checkmark$ & $\times$ & 64.1 & 0.27\\
        PirlNav (IL$\rightarrow$RL)~\cite{ramrakhya2023} & $\times$ & $\checkmark$ & $\times$ & 70.4 & 0.34 \\
        XGX~\cite{Wasserman2023ExploitationGuidedEF} & $\checkmark$ & $\checkmark$ & $\times$ & 72.9 & 0.36\\
        RRR~\cite{mopa} & $\checkmark$ & $\checkmark$ & $\times$ & 30.0 & 0.14\\
        Uni-NaVid~\cite{Zhang2025} & $\times$ & $\checkmark$ & $\times$ & 73.7 & 0.37\\
        FiLM-Nav~\cite{yokoyama2025film} & $\checkmark$ & $\checkmark$ & $\times$ & 77.0 & 0.41\\
        \midrule
         \semnav-OS & $\times$ & $\times$ & $\checkmark$ & 72.9 & 0.34 \\
         \semnav-OS+DINO & $\times$ & $\times$ & $\checkmark$ & 73 & 0.34 \\
         \semnav-RGBS & $\times$ & $\checkmark$ & $\checkmark$ & 74.3 & 0.35 \\
         \semnav-RGBS+DINO & $\times$ & $\checkmark$ & $\checkmark$ & \cellcolor{pink!60}76.2 & \cellcolor{pink!60}0.36  \\
         \semnav-RGBS+DINO $\rightarrow$ RL & $\times$ & $\checkmark$ & $\checkmark$ & \cellcolor{pink!60}77.75 & \cellcolor{pink!60}0.40  \\
        \bottomrule
    \end{tabular}
    }
    \caption{Comparison of \semnav approaches with the state-of-the-art models in the \objnav task of the HM3D
    dataset (validation set). We report the performance using the metrics: SR and SPL.}
    \label{table:state_of_the_art}
\end{table}

%Discussion of the results
Table \ref{table:state_of_the_art} includes, to the best of our knowledge, a comparison with the models that define the state of the art in the \objnav problem.
Based on the results, we would like to highlight the following conclusions.
First, it is noteworthy that our \semnav-OS model, the simplest one as it only uses semantic segmentation information from the environment, outperforms all state-of-the-art models except XGX~\cite{Wasserman2023ExploitationGuidedEF} and FiLM-Nav~\cite{yokoyama2025film}.
We believe this underscores the importance of semantic segmentation information for navigation tasks.
Second, when incorporating RGB information as a sensor, our \semnav-RGBS model surpasses all other approaches in terms of SR, setting a new state of the art on the validation set of the HM3D dataset.
Third, if we apply finetuning to our best model using RL (\semnav-RGBS+DINO $\rightarrow$ RL), following the strategy in \cite{ramrakhya2023}, we achieve an additional improvement, bringing the SR to a value above 77\%.
Notably, FiLM-Nav~\cite{yokoyama2025film} achieves a remarkable SPL that exceeds even our RL-finetuned model (\semnav-RGBS+DINO $\rightarrow$ RL); however, FiLM-Nav operates in 2D and is therefore unable to navigate across multiple floors within a building, which limits its applicability in multi-story environments.
And fourth, in the qualitative results, we observed an interesting exploratory capability in the \semnav models.

Egocentric videos recorded during evaluations in the simulation environment show how \semnav models tend to explore the environment, probing the different rooms they encounter.
If a room is not useful for reaching the target object category, they proceed to leave it and continue exploring.
The videos show that \semnav models can traverse hallways, inspect rooms, and move between different floors of a building.
Moreover, they do not exhibit erratic movements during navigation, reflecting efficient and consistent decision-making.
The promising results obtained in the evaluation metrics support these qualitative observations.
It is worth mentioning that these behaviors have also been observed in experiments conducted in the real world (see Section~\ref{sec:real_world_results}).

\begin{figure}[t]
\centering
\subfloat[]{\includegraphics[width=0.48\linewidth]{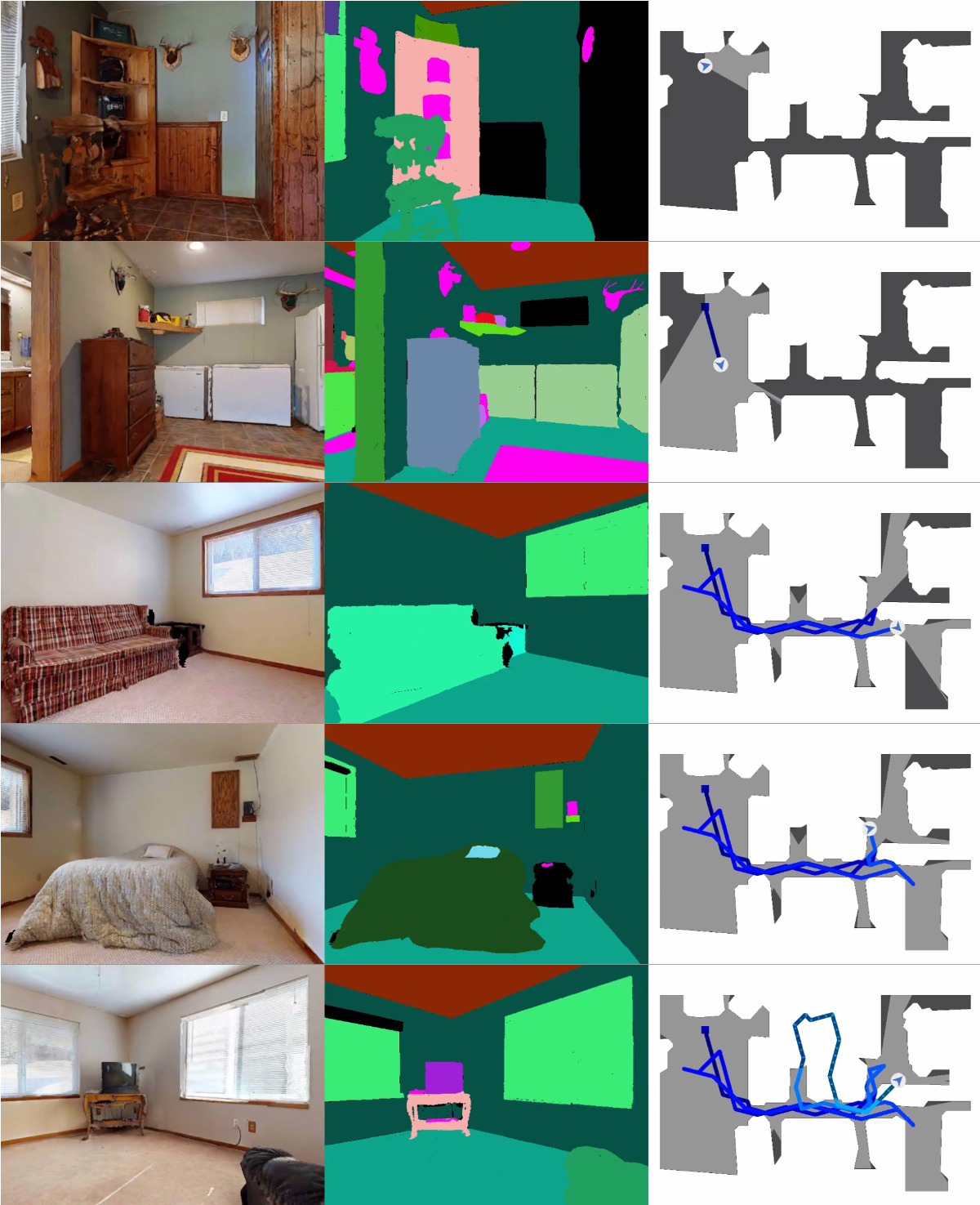}}\hfill
\subfloat[]{\includegraphics[width=0.48\linewidth]{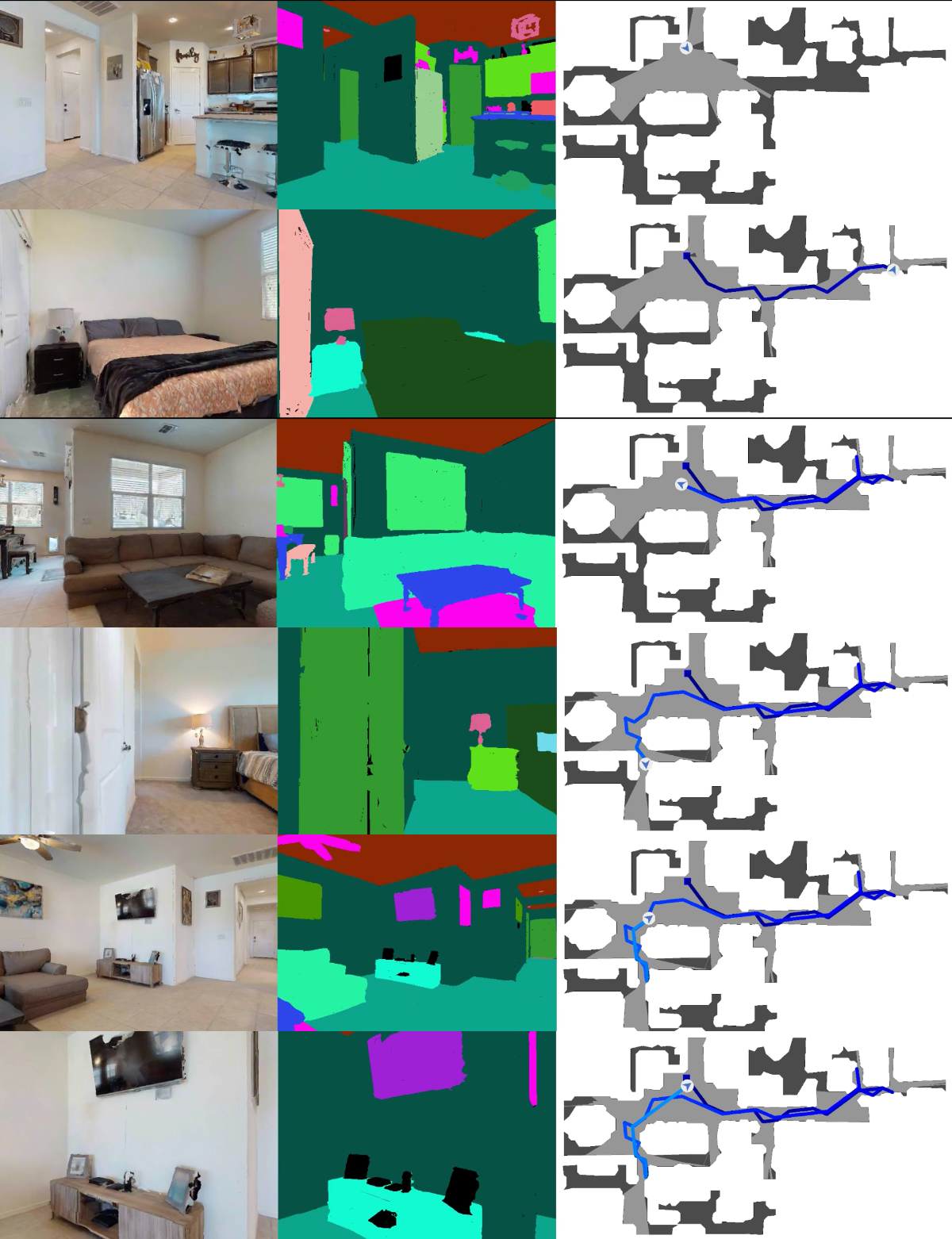}}
\caption{Qualitative results in simulated environments. From top to bottom, these figures show the trajectory followed by the agent in the environment. On the left, the model's input is illustrated as an RGB image; in the center, the semantic segmentation input perceived by the \semnav model; and on the right, the map reflecting the agent's path.}\label{fig:simulation_qualitative}
\end{figure}

Figure \ref{fig:simulation_qualitative} illustrates some of these qualitative navigation results.
For example, Figure \ref{fig:simulation_qualitative} (a) shows how the agent begins its search in a living room, where it does not find the desired object—in this case, a television screen.
It then explores all the rooms on the current floor.
After failing to locate the object, the agent moves to the upper floor, where it finally identifies the requested object.
Figure \ref{fig:simulation_qualitative} (b) depicts another interesting navigation episode.
Initially, the agent is in a large living room where it cannot detect the requested object category.
In the following images, the agent moves through the environment, passing through two bedrooms and viewing the living room from a different perspective.
Finally, the agent detects a television and executes a stop action next to it.
We provide more qualitative results, including failure cases, in the \href{https://youtu.be/ZSjb7tlV2W4}{following video}.

\subsection{Results in the real world}
\label{sec:real_world_results}

The real-world experiments evaluate the performance of the \semnav models in real environments.
First, a comparative analysis of different models is conducted, examining the differences in their results.
These experiments were carried out in house 1 shown in Figure~\ref{fig:house1}.

For the comparative analysis of model performance, we evaluated the \semnav-OS, \semnav-RGBS, and \semnav-RGBS$\rightarrow$RL versions in the real world, using DINO~\cite{Caron2021EmergingPI} pre-training.
Additionally, to compare with the state of the art, the PirlNav model~\cite{ramrakhya2023}, trained using BC with the same dataset, was also included in the evaluation.
The results obtained by these four models are presented in Table~\ref{table:real_world_sr}.

\begin{figure}
    \centering
    \includegraphics[width=0.6\linewidth]{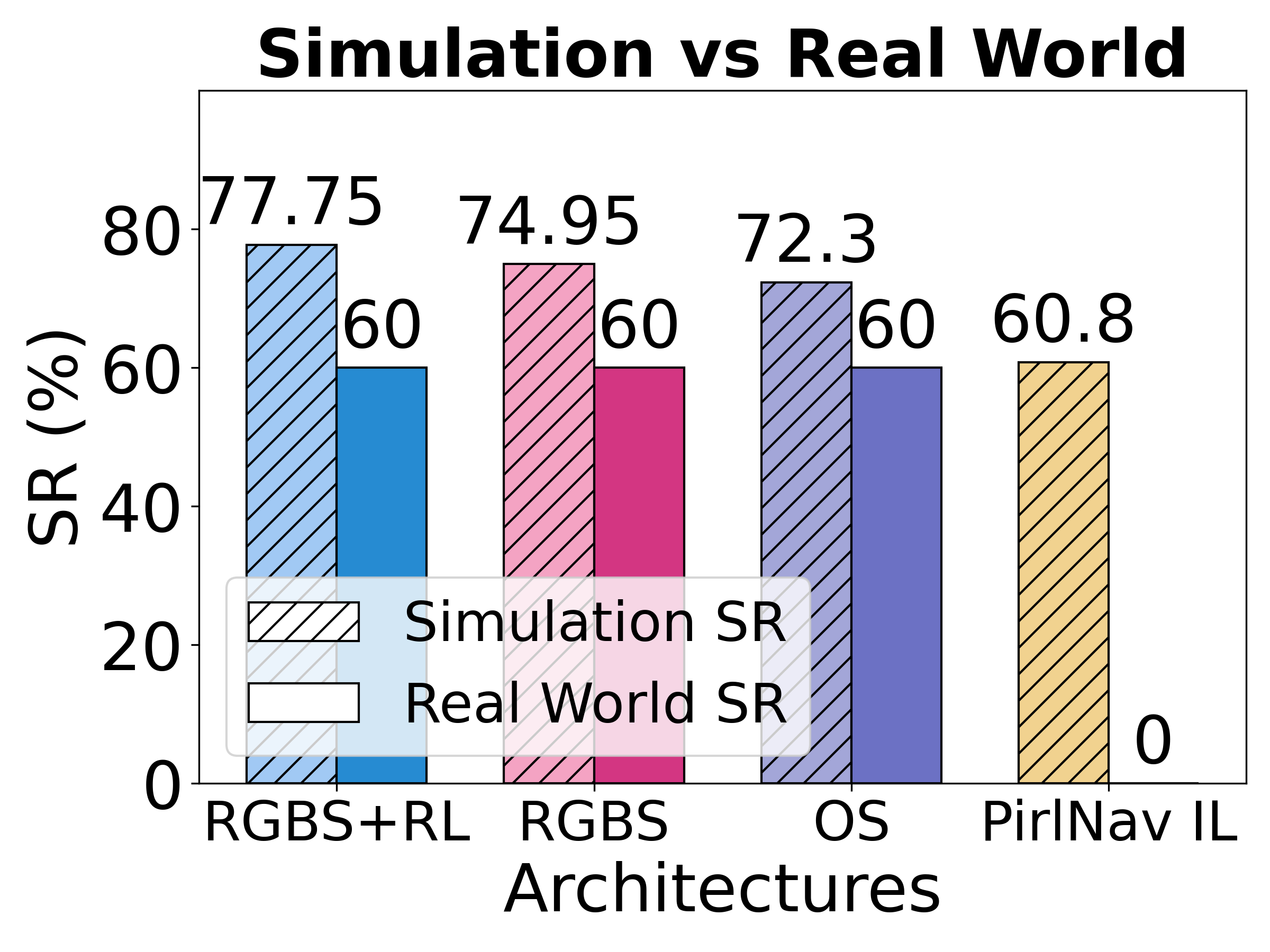}
    \caption{Comparison of the Success Rate (SR) reported in the real world (House 1) and the simulation environment for the different VSN architectures used.}
    \label{fig:real_vs_simulation}
\end{figure}

It is important to note that the PirlNav model did not manage to successfully complete any of the evaluation tasks.
This aspect has already been reported in other works (e.g., \cite{Wasserman2023ExploitationGuidedEF}).
\semnav-OS, which only uses semantic segmentation information, reports the highest SDS, and a lower average number of actions than the \semnav-RGBS, for the same SR.
Only when we use the fine-tuned version with RL, i.~e.~\semnav-RGBS$\rightarrow$RL, the average
number of actions decreases significantly.
Figure \ref{fig:real_vs_simulation} shows a comparison between the results of these four models in the simulation environment and in the real world house 1.
The \semnav-OS model experiences the least gap.

\begin{table}[t]
    \centering
    \resizebox{\linewidth}{!}{%
    \begin{tabular}{lcccccc|c}
        & \multicolumn{6}{c}{\textbf{SR} / \textbf{Actions}} & \textbf{SDS} \\
         \toprule
        & \textbf{Chair} & \textbf{Bed} & \textbf{Toilet} & \textbf{Sofa} & \textbf{TV Monitor} & \textbf{Average} & \\
        \midrule
        PirlNav (RGB)~\cite{ramrakhya2023}    & 0\% / 66  & 0\% / 81  & 0\% / 82  & 0\% / 225  & 0\% / 74  & \textbf{0\%} / 105.6 & 0\\
        \semnav-OS & 100\% / 36 & 0\% / 164  & 0\% / 55  & 100\% / 35 & 100\% / 86 & \textbf{60\%} / 75.2 & 0.019\\
        \semnav-RGBS & 100\% / 40 & 0\% / 71  & 0\% / 140  & 100\% / 49 & 100\% / 94 & \textbf{60\%} / 78.8 & 0.016\\
        \semnav-RGBS$\rightarrow$RL & 100\% / 29 & 0\% / 61  & 100\% / 99  & 100\% / 40 & 0\% / 39 & \textbf{60\%} / 53.6 & 0.018\\
        \bottomrule
    \end{tabular}
    }
    \caption{SR and number of actions per object category, and SDS, reported in real-world experiments conducted in House 1.}
    \label{table:real_world_sr}
\end{table}

\begin{figure}
    \centering
    \includegraphics[width=0.95\linewidth]{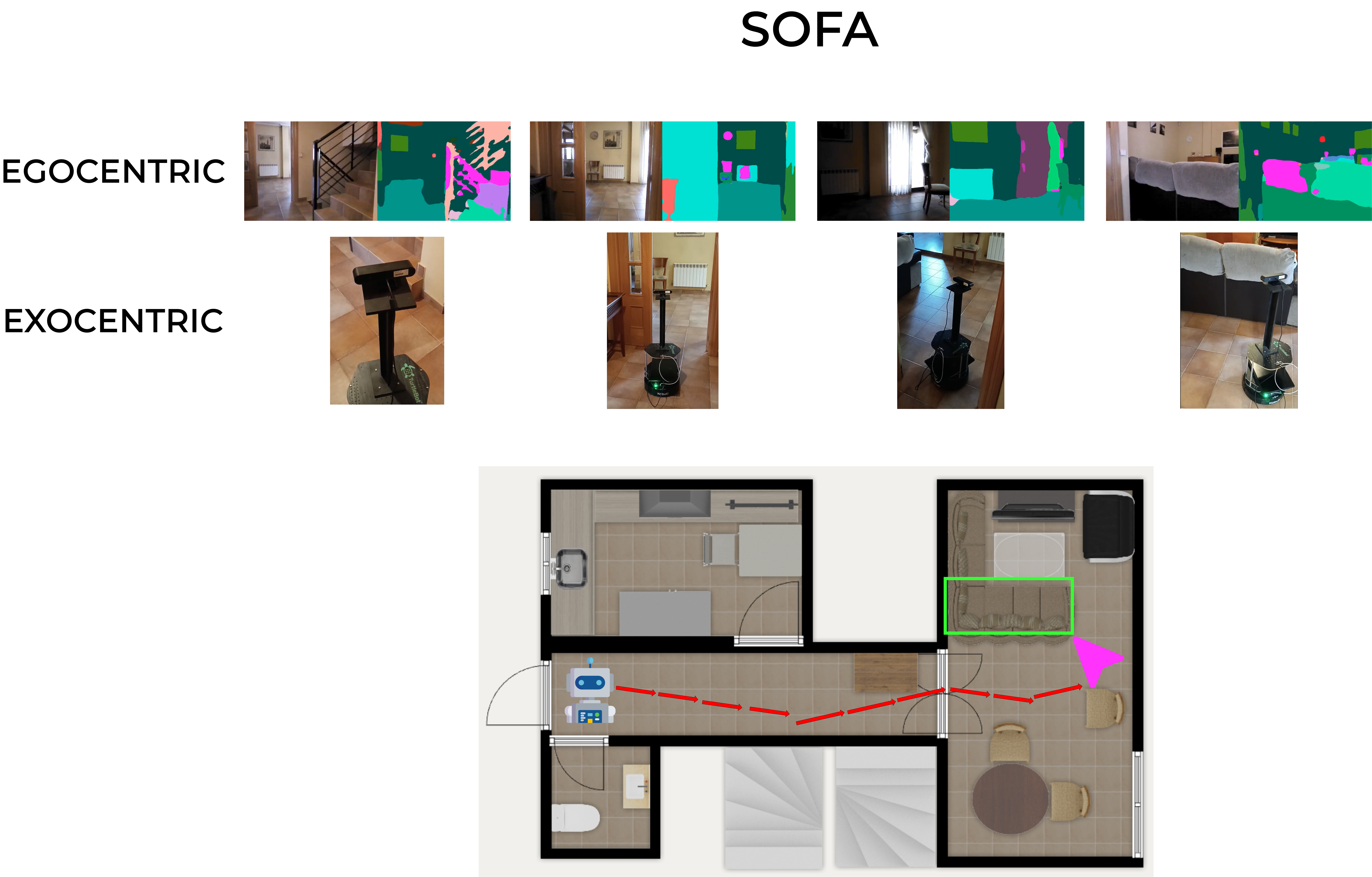} \\
    \includegraphics[width=0.95\linewidth]{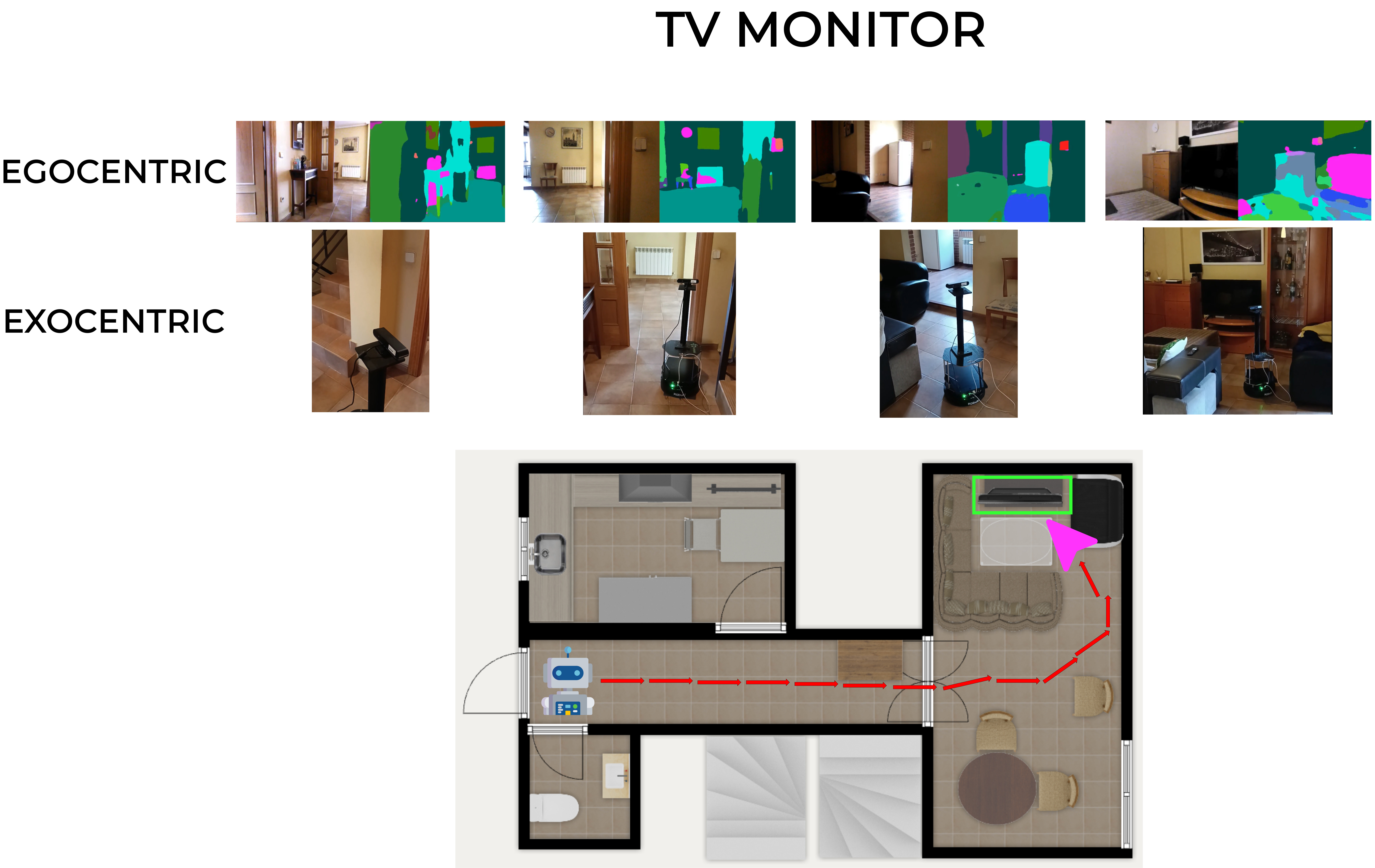} \\
    \includegraphics[width=0.95\linewidth]{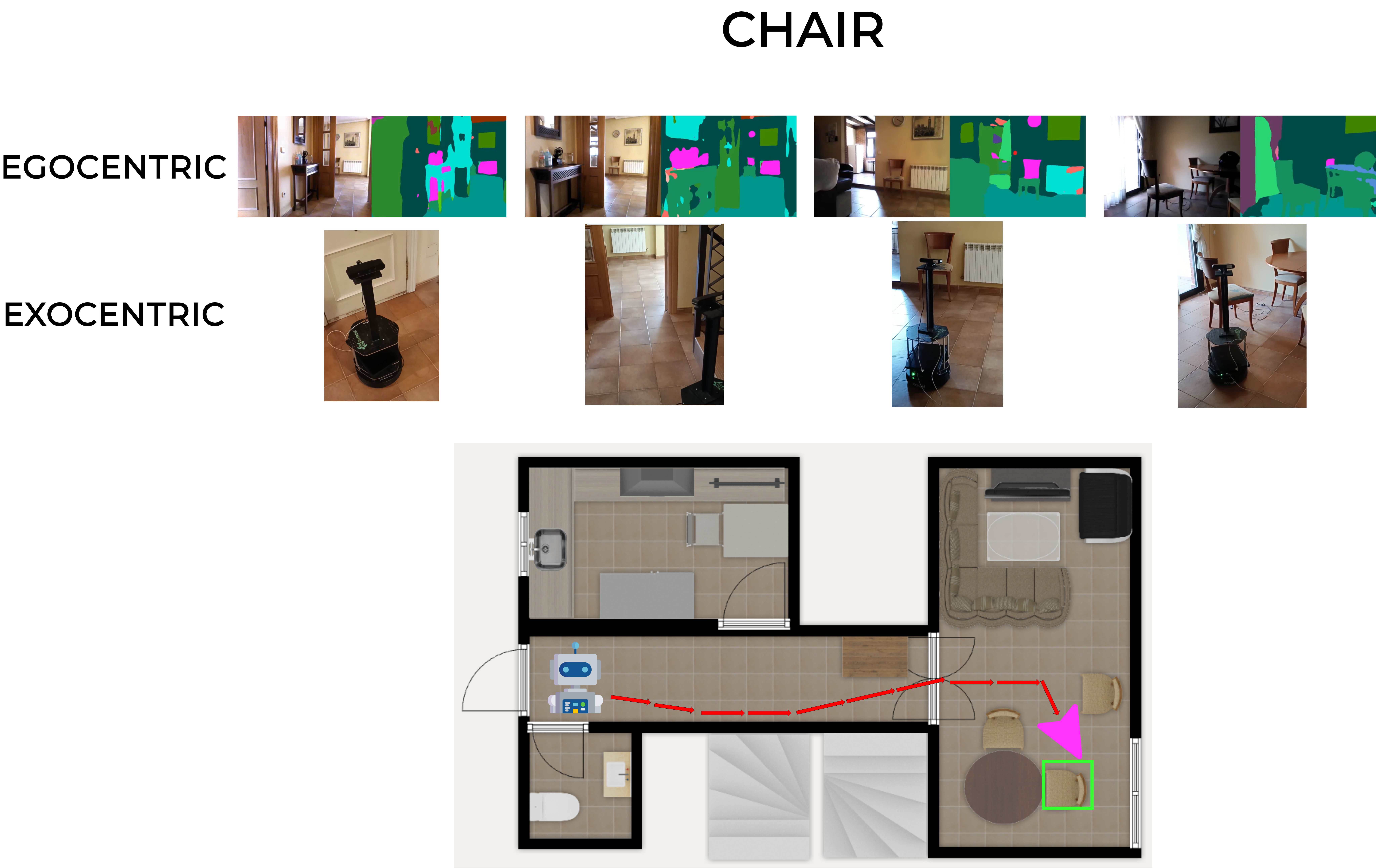}
    \caption{Qualitative results of the robot successfully navigating in the real world House 1 toward a sofa, a television, and a chair.}
    \label{fig:real_world_qualitative}
\end{figure}

None of the models were able to reach the bed object, as it was not present in the navigable scene of house 1.
Interestingly, our \semnav models attempted to climb the stairs when tasked with finding the bed.
This behavior arises because, in the simulated environment, the agents can navigate stairs.
Moreover, beds are typically located on upper floors, which highlights the type of behavior encoded in the learned navigation policy.
Unfortunately, in the real world, our robot lacks this capability.
Figure \ref{fig:real_world_qualitative} presents some qualitative results in house 1, but more can be found in the \href{https://youtu.be/ZSjb7tlV2W4}{following video}, including failure cases.

\begin{figure}[t]
    \centering
    \subfloat[]{\includegraphics[width=0.48\linewidth]{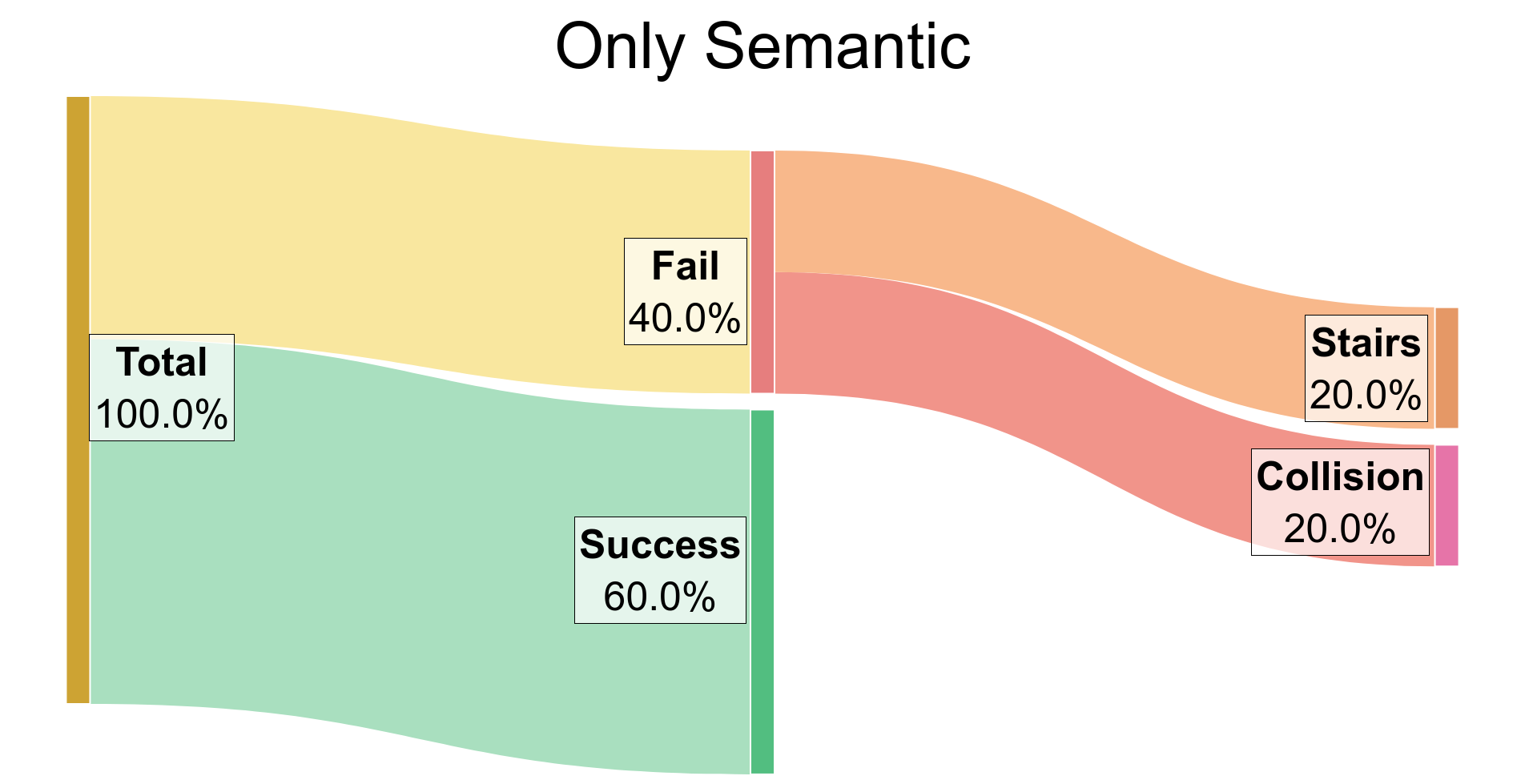}}\hfill
    \subfloat[]{\includegraphics[width=0.48\linewidth]{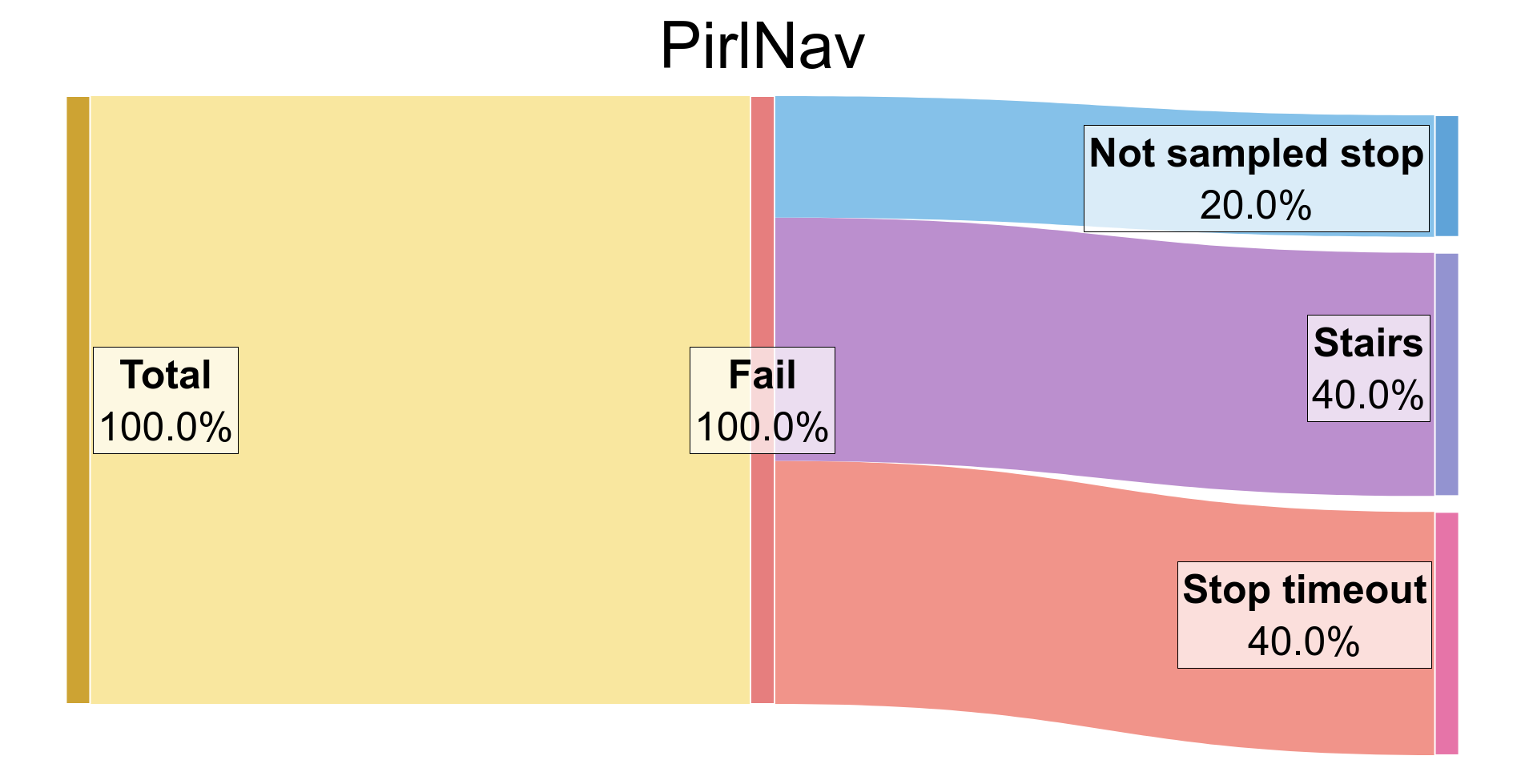}}\\
    \subfloat[]{\includegraphics[width=0.48\linewidth]{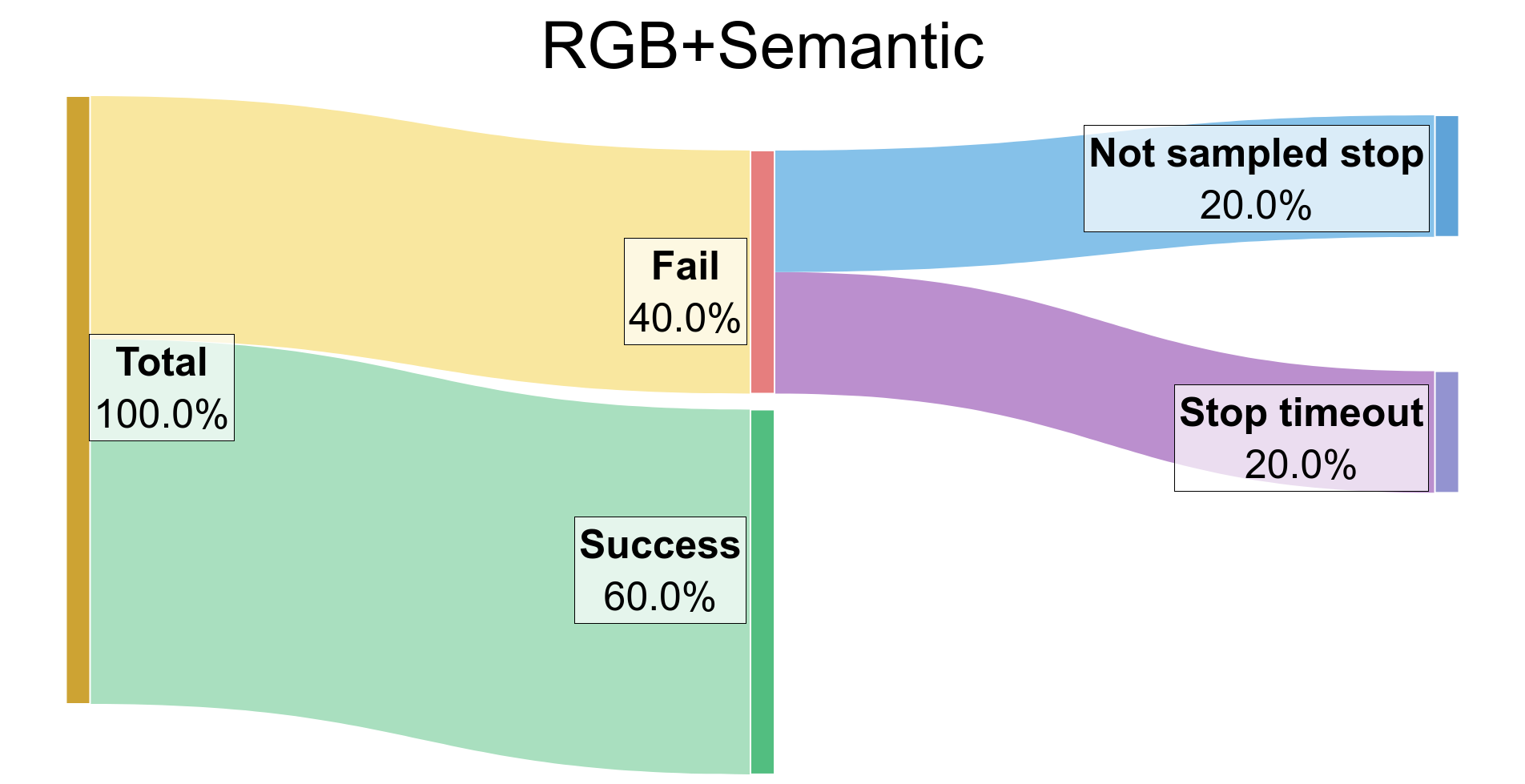}}\hfill
    \subfloat[]{\includegraphics[width=0.48\linewidth]{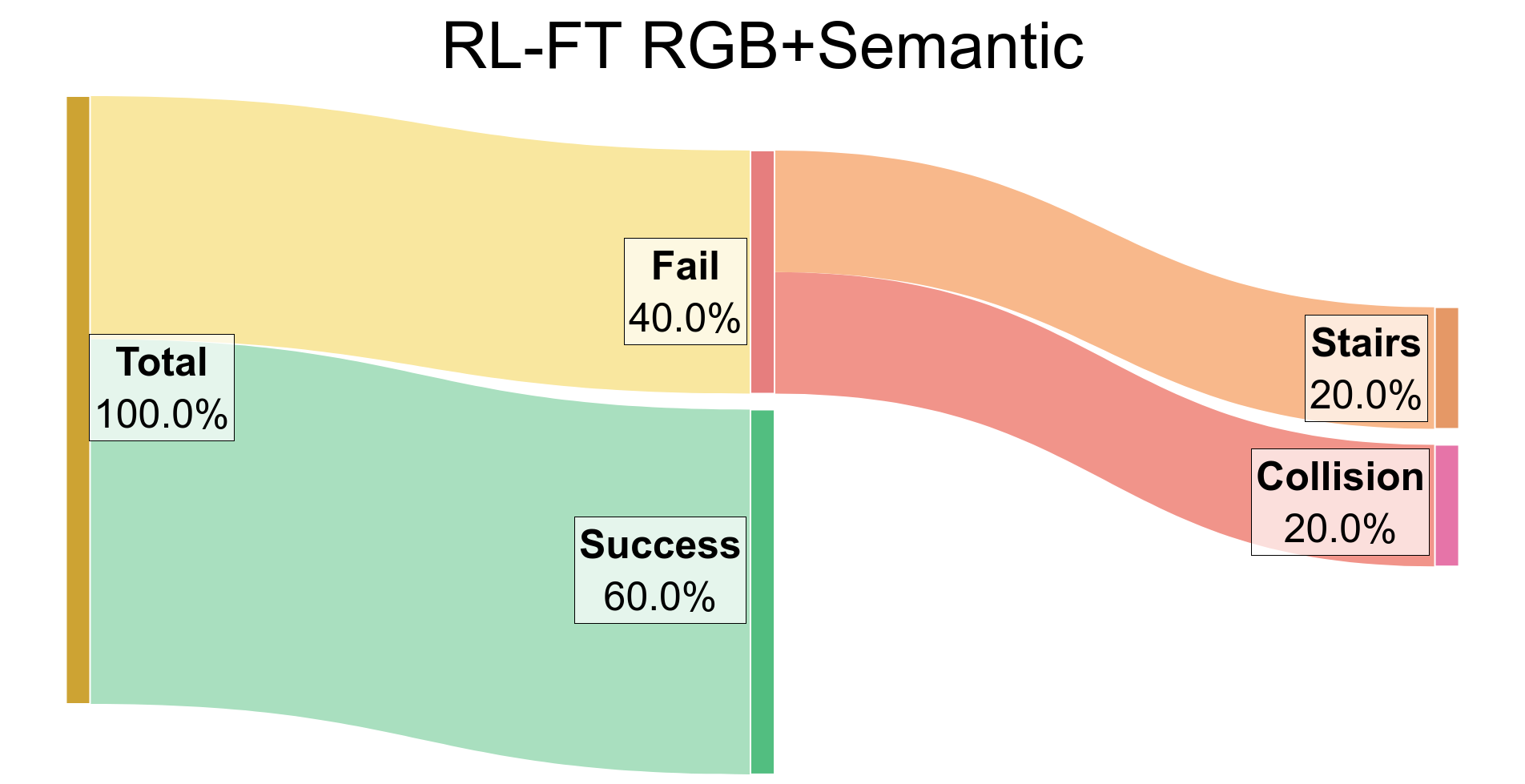}}
    \caption{Failure case analysis for the evaluated models in House 1: (a) \semnav-OS, (b) PirlNav, (c) \semnav-RGBS, and (d) \semnav-RGBS$\rightarrow$RL.}
    \label{fig:failure_analysis}
\end{figure}

Figure~\ref{fig:failure_analysis} presents a detailed analysis of failure cases, comparing the behavior of different models under identical conditions and within the same scenario.
The PirlNav model, which relies solely on RGB information, exhibits numerous failures primarily attributed to ineffective navigation strategies and the failure to execute the \stopac action when approaching the target object.
In contrast, our semantic segmentation-based models (\semnav-OS, \semnav-RGBS, and \semnav-RGBS\(\rightarrow\)RL) demonstrate successful navigation episodes alongside failures predominantly caused by collisions with objects or stairs—limitations inherent to the robotic platform itself.
These models also exhibit occasional failures due to not sampling the \stopac action or exceeding the maximum number of allowed actions.
Overall, the analysis reveals that semantic segmentation information enables more robust navigation behaviors, with failures stemming more from platform constraints than from inadequate decision-making policies.

Furthermore, to complete the real-world evaluation, additional experiments were conducted using the best-performing model, \semnav-RGBS$\rightarrow$RL, in two additional domestic environments (houses 2 and 3).
The results of these experiments are summarized in Table~\ref{table:real_world_RL_FT}, where House~1, House~2, and House~3 correspond to the floor plans shown in Figures~\ref{fig:house1},~\ref{fig:house2}, and~\ref{fig:house3}, respectively.

Table~\ref{table:real_world_RL_FT} shows that, despite navigating in different houses unseen during training, the robot's performance remains robust, achieving an overall SR of 73.7\%, close to the 77.75\% obtained in simulation.
It is particularly noteworthy that the ``bed'' category presents failures in two different houses, both caused by collisions with a sofa, as the semantic segmentation used exhibited noise, and both categories are frequently confused by the segmenter.
Furthermore, this model exhibits a relatively low number of steps, resulting in more efficient navigation trajectories.

\begin{table}[t]
    \centering
    \resizebox{\linewidth}{!}{%
    \begin{tabular}{lcccccc|c}
        & \multicolumn{6}{c}{\textbf{SR} / \textbf{Actions}} & \textbf{SDS} \\
         \toprule
        & \textbf{Chair} & \textbf{Bed} & \textbf{Toilet} & \textbf{Sofa} & \textbf{TV Monitor} & \textbf{Average} & \\
        \midrule
        House 1   & 100\% / 29 & 0\% / 61  & 100\% / 99  & 100\% / 40 & 0\% / 39 & \textbf{60\%} / 53.6 & 0.018\\
        House 2 & 100\% / 20 & 100\% / 58  & 100\% / 79  & 100\% / 12 & 100\% / 77 & \textbf{100\%} / 49.2 & 0.02\\
        House 3 & 100\% / 21 & 0\% / 53  & 100\% / 40  & 0\% / 63 & 100\% / 54 & \textbf{60\%} / 46.2 & 0.026\\
        \bottomrule
        Average & 100\% / 23.3 & 33.3\% / 57.3  & 100\% / 72.7  & 66.7\% / 38.3 & 66.7\% / 56.7 & \textbf{73.3\%} / 49.66 & 0.024\\
        \bottomrule
    \end{tabular}
    }
    \caption{SR, number of actions and SDS obtained in real-world experiments using the \semnav-RGBS$\rightarrow$RL model across three different scenarios with varying lighting conditions. The last row reports the average performance over all experiments.}
    \label{table:real_world_RL_FT}
\end{table}

\begin{figure}
    \centering
    \subfloat[Navigation toward a bed in House~2]{\includegraphics[width=0.48\linewidth]{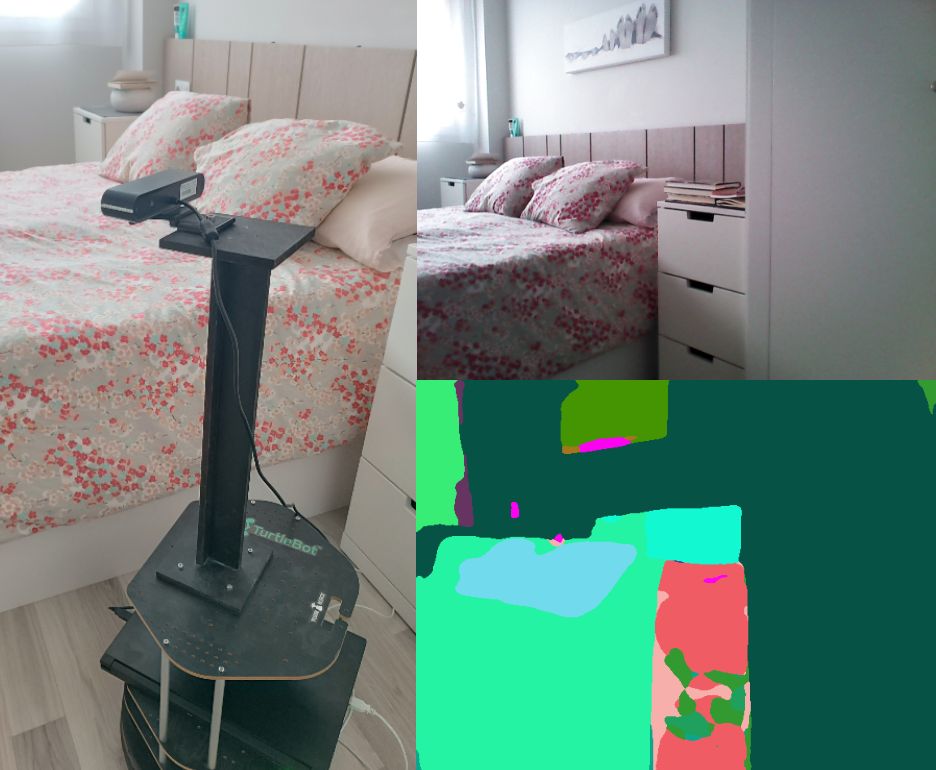}\label{fig:real_world_qualitative_rlft_a}}\hfill
    \subfloat[Navigation toward a TV monitor in House~3]{\includegraphics[width=0.48\linewidth]{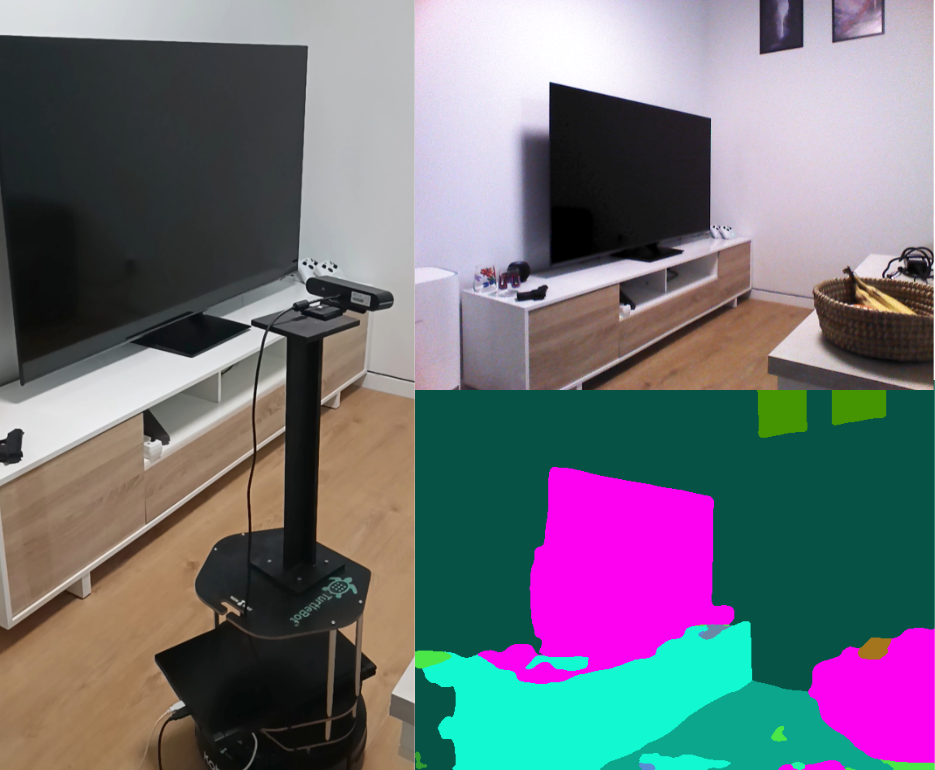}\label{fig:real_world_qualitative_rlft_b}}
    \caption{Qualitative results of the robot successfully navigating in the real world toward a bed in House~2 (left) and toward a TV monitor in House~3 (right), corresponding to the floor plans in Figures~\ref{fig:house2} and~\ref{fig:house3}, respectively.}
    \label{fig:real_world_qualitative_rlft}
\end{figure}

Figure~\ref{fig:real_world_qualitative_rlft} presents qualitative navigation results in House~2 and House~3.
These experiments demonstrate that the agent is capable of successfully navigating toward its target object across different environments, with varying furniture layouts and under different lighting conditions, by leveraging both RGB and semantic segmentation information.

In these real-world experiments, conducted on a TurtleBot 2 platform equipped with an onboard computer featuring an Intel Core i7-10750H CPU @ 2.60 GHz (12 cores) and no dedicated GPU hardware, all models were executed exclusively on the CPU.
This setup accurately reflects the computational constraints of typical robotic platforms employed in practical field applications (where no powerful GPU is embedded in the robot).
Table~\ref{tab:cpu_times} reports the inference times per step obtained under this configuration.
As expected, CPU-based inference leads to a substantial increase in computational latency compared to GPU execution, particularly for the semantic segmentation model.
The action decision models (\semnav-OS and \semnav-RGBS) demonstrate relatively lower inference times, yet they remain significantly higher than their GPU-based counterparts.
The overall per-step inference time—encompassing both perception and decision stages—constitutes a key limitation for achieving real-time navigation performance in resource-constrained robotic systems.
However, these times can be improved by embedding a GPU in the robot, or even using a distributed ROS architecture. For example, the ROS node in charge of semantic segmentation and navigation decisions could have a powerful GPU.

\begin{table}[ht]
    \centering
    \caption{Temporal breakdown per inference step on CPU (ms). ESANet provides semantic segmentation; \semnav-OS and \semnav-RGBS are action decision models. ``Total'' = action model + ESANet.}
    \label{tab:cpu_times}
    \resizebox{\linewidth}{!}{%
    \begin{tabular}{lrrrr}
        \toprule
        \textbf{Model} & \textbf{Inference Time (ms)} & \textbf{+ ESANet (ms)} & \textbf{Total (ms)} \\
        \midrule
        \semnav-OS & 79 & 1400 & 323 \\
        \semnav-RGBS & 137 & 1400 & 387 \\
        \bottomrule
    \end{tabular}
    }
\end{table}

We conclude that although our \semnav models did not always successfully reach the target category, their navigation exhibited a clear intention to move toward it.
The robotic platform’s inability to traverse stairs penalized our models.
Finally, in real-world experiments, the \semnav models demonstrated superior performance compared to the other state-of-the-art models.
This suggests that semantic segmentation helps mitigate the domain gap between real and simulated environments, enabling more accurate navigation in real-world scenarios too.

\section{Conclusion}
%Outline the main contributions
In this paper, we introduced \semnav, a novel VSN model that integrates semantic segmentation as the main visual input to improve navigation efficiency and generalization in unknown environments.
Unlike conventional VSN methods that rely on raw RGB images and struggle with sim-to-real transfer, \semnav benefits from the structured semantic segmentations representations, enabling more robust decision-making in both simulated and real-world environments.
To support our model, we have released the \semnav dataset, designed for training semantic segmentation-aware VSN models, enabling further research in this direction.

Our extensive evaluation shows that \semnav outperforms state-of-the-art \objnav models in both the Habitat 2.0 simulator and real-world tests.
By leveraging semantic segmentation priors, our model achieves higher SR and SPL,
even in
unseen environments, and exhibits enhanced adaptability in real-world scenarios.
The integration of semantic segmentation also reduces the domain gap, a persistent challenge in VSN research.

%Limitations
Despite \semnav’s strong performance in simulation and real-world scenarios, its reliance on semantic segmentation data may limit its applicability in environments where pre-trained models are ineffective, such as forests or underwater settings, due to dataset scarcity and labeling challenges.
Expanding its applicability to other domains may require labor-intensive annotation efforts, as getting additional semantic segmentation datasets remains a significant challenge, although weakly-supervised strategies could be explored.

Experimental results show that performance depends on the semantic segmentation step.
Models that introduce higher segmentation noise perform worse in simulation than those with lower or no noise, indicating that segmentation quality is a critical factor for \semnav.
Obtaining high‑quality semantic segmentations in real environments remains an open challenge, especially in complex or dynamic scenes.
Given this challenge, using higher-quality semantic segmentations could further improve \semnav's performance in real-world scenarios and further reduce the sim-to-real domain gap.
In addition, generating these segmentations incurs substantial computational latency, which can limit applicability in real‑time robotic scenarios.

%Future work
Despite the improved real-world results, real-world navigation necessarily requires a social component that is not addressed in this work.
In future work, we plan to integrate social navigation models into the SemNav model to improve interaction with humans and navigation in populated environments.
Additionally, future work will explore integrating additional semantic information, such as natural language descriptions or spatial relationships between objects, to enrich the environment representation and improve the agent's decision-making.

\bibliographystyle{IEEEtran}
\bibliography{main}

\end{document}